\documentclass[lettersize,journal]{IEEEtran}
\usepackage[T1]{fontenc}
\usepackage{amsmath, bm, booktabs,adjustbox}
\usepackage{algorithm}
\usepackage{algorithmicx}
\usepackage{algpseudocode}
\usepackage{graphicx}
\usepackage{amsfonts}
\usepackage{epstopdf}
\usepackage{caption}
\usepackage{multirow}
\usepackage{float}
\usepackage{color}
\usepackage{dsfont}
\usepackage{mathrsfs}
\usepackage{subfigure}
\usepackage{cases}
\usepackage{amssymb}
\usepackage{url}
\usepackage{cite}
\usepackage{hyperref}
\usepackage{epsfig}
\usepackage{stfloats}
\usepackage{tabularx}
\usepackage{lipsum}
\usepackage{makecell}
\usepackage{pifont}
\usepackage[table]{xcolor} 
\usepackage{arydshln} 
\usepackage{array}        

\hypersetup{hypertex=true,
colorlinks=true,
linkcolor=blue,
anchorcolor=blue,
citecolor=blue}
\newtheorem{theorem}{\textbf{Theorem}}
\newtheorem{lemma}{\textbf{Lemma}}

\newtheorem{corollary}{Corollary}

\begin{document}

\title{A Unified Empirical Risk Minimization Framework for Flexible N-Tuples Weak Supervision}
\author{Shuying Huang, Junpeng Li, \emph{Member}, \emph{IEEE}, Changchun Hua, \emph{Fellow}, \emph{IEEE} and Yana Yang \emph{Member}, \emph{IEEE} \thanks{ S. Huang, J. Li, C. Hua, and Y. Yang are with the Engineering Research Center of the Ministry of Education for Intelligent Control System and Intelligent Equipment, Yanshan University, Qinhuangdao, China(hsy0403@foxmail.com;jpl@ysu.edu.cn;cch@ysu.edu.cn;yyn@ysu.edu.cn).}}
\date{}
\maketitle

\begin{abstract}
To alleviate the annotation burden  in supervised learning, N-tuples learning has recently emerged as a powerful weakly-supervised method. While existing N-tuples learning approaches extend  pairwise learning to higher-order comparisons and accommodate various real-world scenarios, they often rely on task-specific designs and lack a unified theoretical foundation. In this paper, we propose a general N-tuples learning framework based on empirical risk minimization, which systematically integrates pointwise unlabeled data to enhance learning performance. This paper first unifies  the data generation processes of N-tuples and pointwise unlabeled data under a shared probabilistic formulation. Based on this unified view, we derive an unbiased empirical risk estimator that generalizes a broad class of existing N-tuples models. We further establish a generalization error bound for theoretical support.  To demonstrate the flexibility of the framework, we instantiate it in four representative weakly supervised scenarios, each recoverable as a special case of our general model.  Additionally, to address overfitting issues arising from negative risk terms, we adopt correction functions to adjust the empirical risk. Extensive experiments on benchmark datasets validate the effectiveness of the proposed framework and demonstrate that leveraging pointwise unlabeled data consistently improves generalization across various N-tuples learning tasks.
\end{abstract}
\begin{IEEEkeywords}
Weakly-supervised learning,  N-tuples learning, pointwise unlabeled data , unbiased risk estimator. 
\end{IEEEkeywords}

\section{Introduction}\label{sec1}
Weakly-supervised learning (WSL)\cite{zhou2018brief} has emerged as a pivotal paradigm for reducing the cost of manual labeling by exploiting supervision signals that are inaccurate \cite{9765651}, incomplete \cite{9361098}, or inexact \cite{cao2021learning}.  Classical WSL scenarios include positive-unlabeled learning (PU learning)\cite{du2015convex,du2014analysis,sakai2018semi}, where only positive and unlabeled instances are available; partial-label learning\cite{9354590, feng2019partial, 0Discriminative}, where each instance is associated with a set of candidate labels, only one of which is correct; and complementary-label learning\cite{feng2020learning, ishida2017learning, ishida2019complementary},  where each label  specifies a class that the instance does not belong to. Other variants include positive-confidence learning \cite{ishida2018binary}, which utilizes unlabeled data accompanied by confidence scores reflecting their likelihood of being positive.  The most challenging framework, however, is the unlabeled-unlabeled (UU) learning \cite{lu2021binary,lu2020mitigating,lu2018minimal}, which constructs classifiers using only unlabeled datasets that differ in class-prior distributions. Collectively, these approaches broaden the applicability of machine learning to complex tasks without exhaustive annotation.

Pairwise weak supervision has also attracted attention for capturing relationships between instance pairs. For example, pairwise comparisons (Pcomp) learning\cite{feng2021pointwise,10561568} captures relative preferences by enforcing that one instance is more likely to be positive than the other; similarity and unlabeled (SU) learning\cite{bao2018classification} determines whether two instances belong to the same class, and not-all-negative pairwise($P_{pos}U$)\cite{huang2025learning} assumes that at least one instance in each pair is positive. While these approaches have proven effective in applications such as recommendation and ranking, their binary nature limits their ability to capture higher-order relationships within larger groups of instances.

\begin{figure}[t!]
	\setlength{\belowcaptionskip}{0.01cm} 
	\centerline{\includegraphics[width=0.5\textwidth]{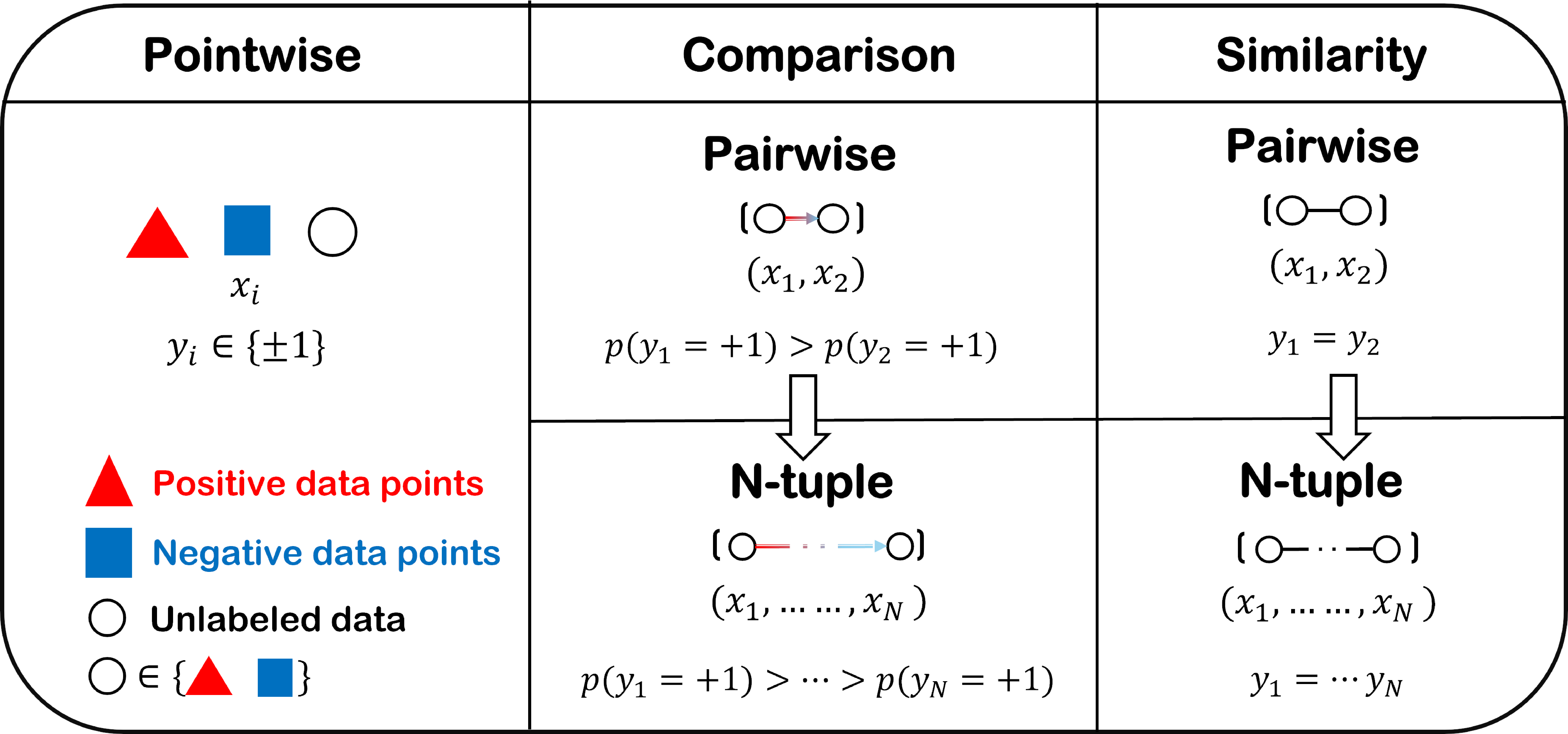}}
	\captionsetup{justification=raggedright,singlelinecheck=false} 
	\caption*{\footnotesize
		The "Comparison" column illustrates Pcomp learning \cite{feng2021pointwise} and NT-Comp learning \cite{LI2025106894}, while the "Similarity" column corresponds to SU learning \cite{shimada2021classification} and NSU learning \cite{0Learning}}
	\caption{Illustration of weak supervision structures from pointwise to pairwise and N-tuple settings.}
	\label{fig:settings}
\end{figure} 
\begin{table}[t]
	\centering
	\caption{Representative weak supervision settings and typical tasks.}
	\label{table:settings}
	\renewcommand{\arraystretch}{1.1}
	\setlength{\tabcolsep}{3pt}
	\begin{tabular}{@{}l p{3.4cm} p{3.9cm}@{}}
		\toprule
		\textbf{Setting} & \textbf{Supervision Assumption} & \textbf{Representative Tasks} \\
		\midrule
		Pointwise & Labels on individual instances (possibly partial or noisy) 
		& PU learning~\cite{du2015convex}, partial label learning~\cite{feng2019partial}, complementary-label learning~\cite{ishida2017learning}, positive-confidence learning~\cite{ishida2018binary}, UU learning~\cite{lu2021binary} \\
		Pairwise  & Constraints on instances pairs 
		& Pairwise comparisons learning\cite{feng2021pointwise}, similarity/dissimilarity learning\cite{shimada2021classification}, and not-all-negative pairwise learning\cite{huang2025learning} \\
		N-tuple   &  Constraints on instances groups 
		& N-tuples comparisons learning \cite{LI2025106894}, N-tuples similarities and unlabeled learning\cite{0Learning}\\
		\bottomrule
	\end{tabular}
\end{table}

\begin{table*}[htbp]
	\centering
	\caption{Mathematical definitions and real-world applications of different N-tuples learning scenarios}
	\label{table:1}
	\small 
	\begin{adjustbox}{width=\textwidth} 
		\begin{tabular}{@{}l >{\centering\arraybackslash}m{5.3cm} >{\centering\arraybackslash}m{6.5cm} >{\centering\arraybackslash}m{2cm}@{}}
			\toprule
			\textbf{Task types} & \textbf{N-tuples scenarios} & \textbf{Mathematical definition} & \textbf{Pointwise unlabeled data} \\
			\midrule
			
			NT-Comp\cite{LI2025106894} & The \(N\) instances are ranked in descending order of confidence for being positive & 
			\(\forall i \in \{1,\dots,N{-}1\},\ \mathbb{P}(y_i = +1) > \mathbb{P}(y_{i+1} = +1)\) & \ding{56} \\
			\midrule
			
			NSU\cite{0Learning} & All \(N\) instances are from the same class & 
			\(\forall i,j \in \{1, \dots, N\},\ y_i = y_j\) & \ding{52} \\
			\midrule
			
			MNU & \textit{Mixed-class N-tuples:} Not all \(N\) instances are from the same class & 
			\(\exists i,j \in \{1, \dots, N\},\ y_i \ne y_j\) & \ding{52} \\
			\midrule
			
			$N_{pos}U$ & \textit{Not-all-negative N-tuples:} At least one instance among the \(N\) is positive & 
			\(\exists i \in \{1, \dots, N\},\ y_i = +1\) & \ding{52} \\
			\bottomrule
		\end{tabular}
	\end{adjustbox}
	\vspace{1mm}
	\caption*{\footnotesize \textbf{Note:} NT-Comp~\cite{LI2025106894} and NSU~\cite{0Learning} are representative existing methods, while MNU and $N_{pos}U$ are novel task settings derived and discussed in this paper under the unified N-tuples learning framework.}
\end{table*}

To address this challenge, recent works have introduced N-tuple weak supervision to handle group-wise relations. For example,
N-tuple comparison learning (NT-Comp)\cite{LI2025106894} assumes a ranking over the 
N instances in each tuple based on their probabilities of being positive, providing richer ordinal constraints among the group. In the N-tuple similarity and unlabeled (NSU)\cite{0Learning} setting, all N instances in a tuple are known to share the same (unknown) label, capturing group-level similarity. Figure~\ref{fig:settings} visually illustrates the structural progression from pointwise to pairwise and N-tuple supervision. While Table~\ref{table:settings} summarizes representative weak supervision paradigms. Compared to pairwise signals, N-tuple supervision provides a more expressive framework for modeling high-order dependencies among multiple instances. However, existing N-tuple methods are typically tailored to specific tasks and lack generalizability across broader settings. 

In this work, we propose a unified N-tuple weakly-supervised learning framework that subsumes existing N-tuple methods and extends them with greater flexibility.  We begin by considering  \(\bar{\mathcal{Y}}\) as the full label space consisting of all \(2^N\) possible label configurations for an N-tuple of binary instances, where each instance is labeled as either positive (\(+1\)) or negative (\(-1\)). For any given weak supervision scenario,  we then define a subset  \(\mathcal{Y}^{\text{sub}} \subseteq \bar{\mathcal{Y}}\) of these assignments that satisfy the task's constraints. 
For example, the NT-Comp scenario corresponds to those assignments where the probabilities of being positive are strictly decreasing across the tuple, i.e., \(\mathbb{P}(y_1 = +1) > \mathbb{P}(y_2 = +1) > \dots > \mathbb{P}(y_N = +1)\), while NSU corresponds to assignments where all labels in the tuple are identical, i.e.,  \(\forall i, j \in \{1, 2, \dots, N\},\ y_i = y_j\). We further introduce two more general scenarios that naturally combine N-tuple weak supervision with pointwise unlabeled data. Specifically, the mixed-class N-tuples and pointwise unlabeled learning  (\textbf{MNU}) allows tuples where not all instances share the same label, i.e., $\exists i, j \in \{1, 2, \dots, N\},\ y_i \neq y_j$; and  the not-all-negative N-tuples and pointwise unlabeled learning ($\boldsymbol{N_{pos}U}$)  ensures at least one positive instance, i.e., \(\exists i \in \{1, 2, \dots, N\},\ y_i = +1\). These settings correspond to practical tasks such as academic performance ranking (NT-Comp), batch quality inspection (NSU), general image classification (MNU), and fraud detection ($N_{pos}U$). By selecting different subsets of the label space, our framework unifies these diverse supervision forms and eliminates the need for task-specific model designs.  The specific label constraints corresponding to each weakly supervised scenario are summarized in Table~\ref{table:1}.

This study builds on the empirical risk minimization (ERM) framework \cite{du2014analysis} and presents a systematic approach to address challenges in weakly-supervised learning. We begin by analyzing the statistical properties of weakly supervision and modeling their underlying distribution. This distribution is then incorporated into the loss function to reconstruct the empirical risk, guiding the model to better leverage weak supervision. In addition,  we provide rigorous theoretical analysis of the proposed framework. By leveraging rademacher complexity, we derive estimation error bounds for both the general formulation and its special cases.  The results show that empirical risk minimization under our framework is statistically consistent: as the number of training instances grows, the learned classifier converges to the best possible classifier under the given constraints. In summary, our method advances both theory and practice by offering a conceptually simple yet powerful framework for N-tuple weak supervision with strong learning guarantees.
The main contributions of this work can be summarized as follows:
\begin{itemize}
   \item \textbf{Unified N-tuple Framework:} We propose a unified framework that models diverse weak supervision scenarios by specifying task-dependent label constraints over the full $2^N$ N-tuple label space. This formulation subsumes existing methods (e.g., NT-Comp, NSU) and naturally generalizes to new settins ( MNU, $N_{pos}U$). The resulting algorithm constructs unbiased risk estimators and jointly optimizes with unlabeled data, offering an efficient and principled learning solution.
  \item \textbf{Theoretical Guarantees:} We establish generalization bounds for both the unified model and its special cases using rademacher complexity. These results confirm the statistical consistency of our approach and provide theoretical insights into learning under weak supervision constraints.
  \item \textbf{Empirical Validation:} We conduct extensive experiments on benchmark datasets across diverse weakly supervised tasks. Our unified method consistently outperforms baseline and specialized models, demonstrating its effectiveness and superior generalizability.
\end{itemize}

\section{Preliminaries}\label{sec2} 
To provide a solid foundation for our research, this section introduces the relevant background of supervised classification method and fundamental concepts.

Supervised classification is a traditional learning paradigm that trains classifiers using precisely labeled examples. Given a dataset with precisely labeled instances, let \(\mathcal{X} \subset \mathbb{R}^d\) denote the feature space consisting of both positive and negative examples.  \(\mathcal{Y} = \{-1,1\}\) indicates the label space of \(\mathcal{X} \), with \(y=1\) denoting a positive instance and \(y=-1\) a negative one. The positive dataset \(\mathcal{X}_{p}\) is independently drawn from the marginal distribution \(p_{+}(\textbf{x}) = p(\textbf{x} \mid y = +1)\). Similarly, the negative sample set \(\mathcal{X}_{n}\) is independently sampled from the marginal distribution \(p_{-}(\textbf{x}) = p(\textbf{x} \mid y = -1)\).  

Thus, each training instance \((\textbf{x}, y) \in (\mathcal{X}, \mathcal{Y})\) is drawn from an unknown joint probability distribution with density \(p(\textbf{x}, y)\). The goal of supervised classification is to learn a classifier \(g: \mathcal{X} \to \mathbb{R}\) by minimizing the expected risk defined as:  
\begin{equation}\label{eq.1}
	\begin{split}
		R(g) = &\mathop{\mathbb{E}}\limits_{p(\textbf{x},y)}[\ell(g(\textbf{x}),y)] \\
		=&\tau_{+}\mathop{\mathbb{E}}\limits_{p_{+}(\textbf{x})}[\ell(g(\textbf{x}),+1)]+
		\tau_{-}\mathop{\mathbb{E}}\limits_{p_{-}(\textbf{x})}[\ell(g(\textbf{x}),-1)].
	\end{split}
\end{equation}
where \(\ell(g(\textbf{x}), y)\) represents the loss function measuring the discrepancy between the classifier's prediction and the true label. Here, $\tau_{+}=p(y=+1)$ represent the class-prior of positive examples and  $\tau_{-}=p(y=-1)$ denote the class-prior of negative examples. These class priors satisfy the constraint $\tau_{+}+\tau_{-}=1$.

Thus, the optimal classifier in supervised classification is obtained by solving:
\begin{equation}\label{eq.2}
g^{*}=\mathop{argmin}\limits_{g\in\mathcal{G}}R(g),
\end{equation}
where \( \mathcal{G} \) denotes the hypothesis space of possible classifiers.
\section{Generalized Framework}\label{sec3}
To accommodate diverse weak supervision settings,  this section proposes a unified framework based on common data generation process. We develop a risk minimization framework aligned with the distributional properties of weakly-supervised data. Estimation error bounds are subsequently established to ensure theoretical guarantees.
\subsection{Generation Process of Training Data}\label{sec3.1}
This section describes the generation process of the training data used in our weakly-supervised learning framework.
 
\bigskip\textbf{N-tuples data :} To standardize the data generation process, we define the sample space  as  \(\bar{\mathcal{D}} = \left\{ \bar{\mathbf{x}}_i \right\}_{i=1}^{\bar{n}} = \left\{ \left( \mathbf{x}_{1,i}, \dots, \mathbf{x}_{N,i} \right) \right\}_{i=1}^{\bar{n}}\), where each \(\bar{\mathbf{x}}_i\)  is an \(N\)-tuple of instances arbitrarily drawn from the feature space, and $\bar{n}$ denotes the total number of such tuples.  
The associated label space is specified as
\begin{equation}\label{eq.10}
 \bar{\mathcal{Y}} =  \{-1,1\}^N.
\end{equation}
The possible $N$-tuple configurations are summarized in Table~\ref{table:2}.

A subset \(\mathcal{Y}^{\text{sub}} \subseteq \bar{\mathcal{Y}}\) is further defined  by imposing specific constraints on the label vectors \(\mathbf{y} = (y_1, \dots, y_N)\), where each \(y_j\in \{-1, 1\}\) denotes the (latent) label of the \(j\)-th instance in the tuple:
\begin{equation}\label{eq.11}
\mathcal{Y}^{\text{sub}} = \left\{ \mathbf{y} \in \bar{\mathcal{Y}} \mid \text{additional constraints} \right\},
\end{equation}
where \(\mathcal{Y}^{\text{sub}} \neq \emptyset\) and \(\mathcal{Y}^{\text{sub}} \neq \bar{\mathcal{Y}}\).

Accordingly, we define a dataset  \(\mathcal{D}_{n} = \left\{ \bar{\mathbf{x}}_i \right\}_{i=1}^{n_{b}},\) consisting of \(N\)-tuples whose latent label vectors belong to  \(\mathcal{Y}^{\text{sub}}\), where $n_{b}$ is the number of such valid tuples.

\begin{lemma}\label{lem:N-dis}
The dataset \(\mathcal{D}_n\) is independently drawn from the distribution \(p_n(\bar{\mathbf{x}})\), given by
\begin{equation}\label{eq.12}
	p_{n}(\bar{\mathbf{x}}) = 
	\frac{
		\sum\limits_{\mathbf{y}\in \mathcal{Y}^{\text{sub}}} 
		\left( \prod\limits_{k=1}\limits^N p_{y_k}(\mathbf{x}_k) \tau_{y_k} \right)
	}{
		\sum\limits_{\mathbf{y} \in \mathcal{Y}^{\text{sub}}} 
		\prod\limits_{k=1}\limits^N \tau_{y_k}
	}.
\end{equation}
where \(p_{y_k}(\mathbf{x}_k)\) denotes the class-conditional density, $\tau_{y_k}=\tau_{+}$ and $p_{y_k} = p_{+}$ if $y_k=+1$; otherwise,  $\tau_{y_k}=\tau_{-}$ and $p_{y_k} = p_{-}$ if $y_k=-1$.
\end{lemma}

The proof is provided in Appendix~A.

\begin{table}[t!]
	\caption{Categorization of N-tuple configurations based on weak supervision types}
	\label{table:2}
	\centering
	\renewcommand\arraystretch{1.2}
	\setlength{\tabcolsep}{3pt}
	\begin{tabularx}{0.5\textwidth}{>{\centering\arraybackslash}X 
			|>{\centering\arraybackslash}X 
			|>{\centering\arraybackslash}X}
		\hline
		Problem & Cases & Description \\
		\hline
		\makecell{Containing $n$ positive \\ $\circledR$ $\circledS$} 
		& \makecell{(+1,+1,+1...+1,+1,+1)\textreferencemark}
		& One case \\ 
		\hline
		\multirow{5}{=}{\makecell[c]{Containing $n\!-\!1$ positive\\ $\circledR$ $\star$}} 
		& (+1,+1,+1...+1, +1,-1)\textreferencemark & \multirow{5}{*}{$\binom{n}{1}$} \\
		& (+1,+1,+1...+1,-1,+1) & \\
		& $\vdots$ & \\
		& (+1,-1,+1...+1,+1,+1) & \\
		& (-1,+1,+1...+1,+1,+1) & \\
		\hline
		\multirow{5}{=}{\makecell[c]{Containing $n\!-\!2$ positive\\ $\circledR$ $\star$}} 
		& (+1,+1,+1...+1,-1,-1)\textreferencemark & \multirow{5}{*}{$\binom{n}{2}$} \\
		& (+1,+1,+1...-1,-1,+1) & \\
		& $\vdots$ & \\
		& (+1,-1,-1...+1,+1,+1) & \\
		& (-1,-1,+1...+1,+1,+1) & \\
		\hline
		$\vdots$ & $\vdots$ & $\vdots$ \\
		\hline
		\multirow{5}{=}{\makecell[c]{Containing one positive\\ $\circledR$ $\star$}} 
		& (+1,-1,-1...-1,-1,-1)\textreferencemark & \multirow{5}{*}{$\binom{n}{n-1}$} \\
		& (-1,+1,-1...-1,-1,-1) & \\
		& $\vdots$ & \\
		& (-1,-1,-1...-1,+1,-1) & \\
		& (-1,-1,-1...-1,-1,+1) & \\
		\hline
		\makecell{Containing $n$ negative\\ $\circledS$}
		& (-1,-1,-1,...-1,-1,-1)\textreferencemark 
		& One case \\ 
		\hline
	\end{tabularx}
	{\footnotesize 
		\begin{flushleft}
			\textreferencemark: N-tuple comparison data (combinations ordered by decreasing confidence of being positive);\\
			$\circledR$: Not-all-negative N-tuples (at least one positive instance);\\
			$\circledS$: Similar N-tuples (all instances from the same class);\\
			$\star$: Mixed-class N-tuples (instances not all from the same class).
		\end{flushleft}
	}
\end{table}

Let $\mathcal{D}_{j} = \left\{ \mathbf{x}_{j,i} \right\}_{i=1}^{n_b}$ denote the dataset of the $j$-th elements extracted from all tuples in $\mathcal{D}_n$,  treating each instance independently and disregarding the original tuple structure. Based on this, we present Theorem~1.
\begin{theorem}\label{the:N-dis} The dataset \(\mathcal{D}_{j}\) is independently sampled from an underlying distribution \(\tilde{p}_j(\mathbf{x})\).
\begin{equation}\label{eq.13}
	\tilde{p}_{j}(\mathbf{x}) = 
	\underbrace{
		\frac{
			\sum\limits_{\substack{\mathbf{y} \in \mathcal{Y}^{\text{sub}} \\ y_j = +1}} \prod\limits_{k=1}^N \tau_{y_k}
		}{
			Z
		}
	}_{\alpha_j} p_{+}(\mathbf{x})
	+
	\underbrace{
		\frac{
		\sum\limits_{\substack{\mathbf{y} \in \mathcal{Y}^{\text{sub}} \\ y_j = -1}} \prod\limits_{k=1}\limits^N \tau_{y_k}
		}{
			Z
		}
	}_{\beta_j} p_{-}(\mathbf{x}),
\end{equation}
where
\( Z = 	\sum\limits_{\mathbf{y} \in \mathcal{Y}^{\text{sub}}} 
\prod\limits_{k=1}\limits^N \tau_{y_k} \). \(\alpha_j\) and \(\beta_j\) represent the weighting coefficients associated with the positive and negative class distributions, respectively.
\end{theorem} 

The proof is provided in Appendix~B.

In the symmetric case, all positions within the $N$-tuple are statistically equivalent, and thus $\tilde{p}_1(\mathbf{x}) = \cdots = \tilde{p}_N(\mathbf{x})$. This implies that $\alpha_j = \alpha$ and $\beta_j = \beta$ for all $j \in \{1, \dots, N\}$.

Furthermore, we consider a ponitwise unlabeled dataset, which plays a crucial role in leveraging the underlying data distribution to improve model generalization under weak supervision.

\bigskip\textbf{Pointwise unlabeled data :} Pointwise unlabeled data  refer to instances that are not associated with any label information, and the corresponding dataset is denoted as   $\mathcal{D}_{u}=\{\textbf{x}_{u,i}\}_{i=1}^{n_{u}}$. Each instance \(\mathbf{x}_{u,i} \in \mathcal{D}_u\) is assumed to be independently drawn from a marginal distribution \(p(\mathbf{x})\), which is expressed as a convex combination of the class-conditional distributions: 
\begin{equation}\label{eq.14}
	p(\textbf{x})=\tau_{+}p_{+}(\textbf{x})+\tau_{-}p_{-}(\textbf{x}).
\end{equation}

To incorporate both N-tuples and pointwise unlabeled data, the following datasets are considered:
\begin{align}
	&\mathcal{D}_n = \left\{ \bar{\mathbf{x}}_i \right\}_{i=1}^{n_b} \sim p_{n}(\bar{\mathbf{x}}),\\
	&\mathcal{D}_{u}=\{\textbf{x}_{u,i}\}_{i=1}^{n_{u}} \sim p(\textbf{x}). 
\end{align}

This formulation integrates structural information from N-tuples and distributional insights from unlabeled instances, thereby laying the foundation for constructing a unified risk function.
\subsection{ Unbiased Risk Estimator for the proposed method}\label{sec3.2}
This section begins by revisiting the generalized linear system that relates the observed mixture distributions to the underlying class-conditional distributions:
\begin{equation}\label{eq.17}
\begin{bmatrix}
	\tilde{\mathbf{p}}(\mathbf{x}) \\
	p(\mathbf{x})
\end{bmatrix}
=
\begin{bmatrix}
	\mathbf{A} \\
	\mathbf{\Gamma}
\end{bmatrix}
\begin{bmatrix}
	p_+(\mathbf{x}) \\
	p_-(\mathbf{x})
\end{bmatrix},
\end{equation}
where \(\tilde{\mathbf{p}}(\mathbf{x}) = [\tilde{p}_1(\mathbf{x}), \tilde{p}_2(\mathbf{x}), \dots, \tilde{p}_N(\mathbf{x})]^\top\) is an \(N\)-dimensional column vector, and the matrix
\(\begin{bmatrix} \mathbf{A} \\ \boldsymbol{\Gamma} \end{bmatrix} \in \mathbb{R}^{(N+1)\times 2}\) aggregates the mixture coefficients.

The matrix \(\mathbf{A} \in \mathbb{R}^{N \times 2}\) is defined as
$$
\mathbf{A} =
\begin{bmatrix}
	\alpha_1 & \beta_1 \\
	\alpha_2 & \beta_2 \\
	\vdots & \vdots \\
	\alpha_N & \beta_N
\end{bmatrix},
\quad \text{and} \quad
\mathbf{\Gamma} = [\tau_+ \quad \tau_-],
$$

In the asymmetric setting, as presented in Section~\ref{sec5.1},  $\mathbf{A}$ is an $N \times 2$ matrix whose coefficients may vary across rows, reflecting heterogeneous instance-level mixtures. In contrast, under the symmetric assumption (Sections~\ref{sec5.2}-\ref{sec5.4}), all rows in $\mathbf{A} = [\alpha\ \beta]$ are identical, implying that all mixtures share the same marginal distribution.

By applying the operations in Eq.~\eqref{eq.17}, Lemma 2 can be derived.

\begin{lemma} \label{lem:con-den}
The class-conditional densities $p_+(\mathbf{x})$ and $p_-(\mathbf{x})$ can be recovered by the following closed-form solution:
\begin{equation}\label{eq.18}
\begin{bmatrix}
	p_+(\mathbf{x}) \\
	p_-(\mathbf{x})
\end{bmatrix}
=
\left(\mathbf{M}^\top \mathbf{M}\right)^{-1} \mathbf{M}^\top 
\begin{bmatrix}
	\tilde{\mathbf{p}}(\mathbf{x}) \\
	p(\mathbf{x})
\end{bmatrix}.
\end{equation}
where $\mathbf{M} = \begin{bmatrix}
	\mathbf{A} \\
	\mathbf{\Gamma}
\end{bmatrix}$. This result holds under the condition that $M^{\top}M$ is invertible, i.e., $M$ has full column rank.
\end{lemma}

The detailed proof of Lemma 2  is provided in Appendix~C.

Supposing 
\begin{equation}\label{eq.19}
	(\mathbf{M}^T \mathbf{M})^{-1} \mathbf{M}^T = \begin{bmatrix}
		C_{11} & C_{12} & \cdots & C_{1N} & D_1 \\
		C_{21} & C_{22} & \cdots & C_{2N} & D_2
	\end{bmatrix},
\end{equation}
Thus, we have,
\begin{equation}\label{eq.20}
	\begin{split}
		p_+(\mathbf{x}) = \sum_{j=1}^N C_{1j} \tilde{p}_j(\mathbf{x}) + D_1 p(\mathbf{x}), \\
		p_-(\mathbf{x}) = \sum_{j=1}^N C_{2j} \tilde{p}_j(\mathbf{x}) + D_2 p(\mathbf{x}).
	\end{split}
\end{equation}

Substituting the expressions for $p_+(\textbf{x})$ and $p_-(\textbf{x})$ into the risk function in Eq.~\eqref{eq.1}, we obtain:

\begin{theorem}\label{thm:asy_general_risk}
The risk function in Eq.~\eqref{eq.1} can thus be rewritten as:
\begin{equation}\label{eq.21}
	\begin{split}
		R_n(g) 
		&= \sum_{j=1}^N \mathop{\mathbb{E}}\limits_{\tilde{p}_j(\textbf{x})}[\tau_{+} C_{1j} \ell(g(\textbf{x}),+1) + \tau_{-} C_{2j} \ell(g(\textbf{x}),-1)]\\
		&+ \mathop{\mathbb{E}}\limits_{p(\textbf{x})} \left[ \tau_{+} D_1 \ell(g(\textbf{x}),+1) + \tau_{-} D_2 \ell(g(\textbf{x}),-1) \right],
	\end{split}
\end{equation}
\end{theorem}

The detailed proof of Theorem 2  is provided in Appendix~D.
Accordingly, the empirical version of the risk function based on sample means is given by:
\begin{equation}\label{eq.22}
	\begin{split}
		&\widehat{R}_{n}(g) \\
		&= \frac{1}{n_b}\sum_{j=1}^N \sum_{i=1}^{n_b} \left[ \tau_{+} C_{1j}\ell(g(\textbf{x}_{j,i}),+1) + \tau_- C_{2j}\ell(g(\textbf{x}_{j,i}),-1)\right]\\
		&+ \frac{1}{n_u} \sum_{i=1}^{n_u} \left[ \tau_+ D_1\ell(g(\textbf{x}_{u,i}),+1) + \tau_-D_2\ell(g(\textbf{x}_{u,i}),-1)\right].
	\end{split}
\end{equation}

The classifier is thus trained by minimizing the empirical risk \(\widehat{R}_{n}(g)\).
\begin{equation}\label{eq.23}
	\hat{g}_n=\mathop{argmin}\limits_{g\in\mathcal{G}}\widehat{R}_{n}(g),
\end{equation}
This setting provides a unified framework to exploit N-tuples structured information and pointwise unlabeled data.

\begin{theorem}\label{thm:sym_general_risk}
If the pointwise data distribution is symmetric, then the risk function simplifies to:
\begin{equation}\label{eq.24}
	\begin{split}
		R_{n}(g) = \frac{\tau_{+}\tau_{-}}{\alpha \tau_{-} - \beta \tau_{+}} \mathop{\mathbb{E}}\limits_{\mathbf{x} \sim \tilde{p}_{j}(\mathbf{x})} \left[ \mathcal{L}_{\ell}(g(\mathbf{x})) \right] 
		+ \mathop{\mathbb{E}}\limits_{\mathbf{x} \sim p(\mathbf{x})} \left[ \mathcal{L}_{u,\ell}(g(\mathbf{x})) \right],
	\end{split}
\end{equation}
\textit{where:}
\begin{align*}
	\mathcal{L}_{\ell}(z) &= \ell(z, +1) - \ell(z, -1), \\
	\mathcal{L}_{u,\ell}(z) &= \frac{\alpha  \tau_{-}\ell(z, -1) - \beta \tau_{+}\ell(z, +1)}{\alpha \tau_{-} - \beta \tau_{+}}.
\end{align*}
\end{theorem}
The detailed proof of Theorem is provided in Appendix~E for completeness.
\subsection{Estimation Error Bound}\label{sec4.3}
This section presents a generalization error bound for the classifier \(\widehat{g}_n\), learned from N-tuples structured data combined with pointwise unlabeled instances.

The derivation relies on the following assumptions regarding the hypothesis class and the loss function.

\textbf{Assumptions:}
\begin{itemize}
 \item Let \(\mathcal{G} \subset \mathbb{R}^{\mathcal{X}}\) be the function class under consideration. Each \(g \in \mathcal{G}\) is uniformly bounded, i.e., \(\|g\|_{\infty} \leq C_g\) for some constant \(C_g > 0\).
 \item The loss function \(\ell\) is \(\rho\)-Lipschitz continuous with respect to its first argument, with \(\rho \in (0, \infty)\). In addition, let \(C_\ell = \sup_{t \in \{\pm1\}} \ell(C_g, t)\) denote the maximum loss value under this bound.
\end{itemize}

To assess the generalization performance of the classifier \(\widehat{g}_n\) trained from N-tuples and pointwise unlabeled data , we derive an estimation error bound based on Rademacher complexity \cite{bartlett2002rademacher}. 

\begin{theorem}\label{thm:asy_general_bound}
 Let
$\widehat{g}_{n} = argmin_{g\in\mathcal{G}} \widehat{R}_{n}(g)$ be the general empirical risk minimizer. For any $\delta > 0$, with probability at least $1-\delta$:
\begin{equation}\label{eq.25}
\begin{split}
&R(\hat{g}_{n})-R(g^{\ast})\leq K_{n}\frac{1}{\sqrt{n_b}}+K_{u}\frac{1}{\sqrt{n_{u}}}
\end{split}
\end{equation}
where \\
$K_{n}=\left(\tau_+ \sum\limits_{j=1}^{N}C_{1j}+\tau_- \sum\limits_{k=1}^{N}C_{2j}\right)\left(4\rho C_{\mathcal{G}}+C_{\ell}\sqrt{2\ln\frac{4N}{\delta}}\right)$ 
and \\
$K_{u}= \left(\tau_+ D_1+\tau_- D_2\right)\left(4\rho C_{\mathcal{G}}+C_{\ell}\sqrt{2\ln\frac{4}{\delta}}\right)$
\end{theorem}

Appendix~F provides the detailed derivation of Theorem 4 and presents Lemma 4, whose proof relies on standard results and can be found, for example, in Theorem 3.1 of \cite{mohri2018foundations}.

Theorem 4 implies that  the learned classifier \(\hat{g}_{n}\) converges to the optimal classifier \(g^{*}\) at the optimal rate of \(\mathcal{O}(\frac{1}{\sqrt{n_b}} + \frac{1}{\sqrt{n_u}})\) as \(n_b \to \infty\) and \(n_u \to \infty\). This result confirms the consistency of the proposed method and highlights its sample efficiency when leveraging both N-tuples structured data and unlabeled instances. 

The error bound under the symmetric condition is formally provided in Theorem 5.

\begin{theorem}\label{thm:sym_general_bound}
For any $\delta > 0$, with probability at least $1 - \delta$, the risk under the symmetric data distribution satisfies the following error bound:
\begin{equation}\label{eq.26}
\begin{split}
	&R(\hat{g}_{n})-R(g^{\ast})\leq S_{n}\frac{1}{\sqrt{ Nn_{b}}}+S_{u}\frac{1}{\sqrt{n_{u}}}
\end{split}
\end{equation}
where $S_{n}=\frac{4\tau_{+}\tau_{-}}{\alpha\tau_{-}-\beta\tau_{+}}(2\rho C_{\mathcal{G}}+C_{\ell}\sqrt{\frac{1}{2}\ln\frac{4}{\delta}})$
and $S_{u}=4\rho C_{\mathcal{G}}+2C_{\ell}\sqrt{\frac{\ln\frac{4}{\delta}}{2}}$.
\end{theorem} 
A supplementary proof of Theorem 5 can be found in Appendix~G.
\begin{algorithm}[t!]
	\caption{Generalized algorithm}
	\label{alg1}
	\begin{algorithmic}[1]
		\renewcommand{\algorithmicrequire}{\textbf{Input:}}
		\Require 
		Model $g$,  N-tuple $\mathcal{D}_n = \left\{ \left( \bar{\mathbf{x}}_i \right) \right\}_{i=1}^{n_b}$ (sampled from $p_{n}(\bar{\textbf{x}})$), Pointwise unlabeled data  $\mathcal{D}_{u}=\{\textbf{x}_{u,i}\}_{i=1}^{n_{u}}$ (sampled from $p(\textbf{x})$);
		\For{$i = 1,2,$ ... number of epochs}
		\State \textbf{Shuffle}  $\mathcal{D}_{c}=\mathcal{D}_{n}\cup\mathcal{D}_{u}$
		\For{$j = 1,2,$ ... number of batch\_size}
		\State \textbf{Fetch} mini-batch $\tilde{\mathcal{D}_{c}}$ from $\mathcal{D}_{c}$
		\State \textbf{Update} model $g$ by minimize risk loss $\widehat{R}_{n}(g)$ in Eq.(23)
		\EndFor
		\EndFor
		\Ensure
		$g$.				
	\end{algorithmic}
\end{algorithm}
\bigskip\section{Bridging generalized N-tuples learning with specific applications}\label{sec5}
This section analyzes the structure of the general model as it specializes to various weak supervision settings. The corresponding supervision scenarios and their label structures are summarized in Table~\ref{table:2}, with distinct symbols indicating different weak supervision types.
\subsection{Case 1: N-tuples comparisons and unlabeled learning ($N_{\text{Comp}}U$)}\label{sec5.1}
To align with the general modeling framework, this section integrates the NT-Comp learning\cite{LI2025106894} by incorporating pointwise unlabeled data  to enhance classification. 

The \textit{N-tuples comparisons data}\cite{LI2025106894} considers a scenario in which the instances within each tuples are ordered according to their confidence of belonging to the positive class. Specifically, for an \(N\)-tuples \((\textbf{x}_1, \textbf{x}_2, \dots, \textbf{x}_N)\), the confidence satisfy a descending order:
\[
\text{conf}(\textbf{x}_1) \geq \text{conf}(\textbf{x}_2) \geq \dots \geq \text{conf}(\textbf{x}_N),
\]
where \(\text{conf}(\textbf{x}_j)\) denotes the confidence of \(\textbf{x}_j\) being positive for \(j = 1, \dots, N\). 

 Accordingly,  the label space of N-tuples comparisons data is defined as:
\begin{equation}\label{eq.3}
	\begin{split}
	\mathcal{Y}^{\text{sub}} &= \mathcal{Y}^{\text{comp}} \\
	&= \{\mathbf{y} \in \{-1,1\}^N \mid \mathbb{P}(y_1 = +1) >  \dots > \mathbb{P}(y_N = +1) \}.
	\end{split}
\end{equation}
This label space encodes all label assignments that are consistent with the descending confidence assumption.

The full N-tuples comparisons dataset is defined as \(\mathcal{D}_c = \left\{ (\mathbf{x}_{c_1,i}, \dots, \mathbf{x}_{c_N,i}) \right\}_{i=1}^{n_c}\), where each tuple contains \(N\) instances ordered by confidence. For pointwise analysis, we extract position-specific instance sets \(\widetilde{\mathcal{D}}_{c_{j}} = \left\{ \mathbf{x}_{c_j,i} \right\}_{i=1}^{n_c}\), where each \(\widetilde{\mathcal{D}}_{c_{j}}\) contains all instances at the \(j\)-th position across tuples. \(\widetilde{\mathcal{D}}_{c_{j}}\) is independently sampled from:
\begin{equation}\label{eq.4}
	\tilde{p}_j(\mathbf{x}) =\alpha_{j} p_+(\mathbf{x}) + \beta_{j}p_-(\mathbf{x}),
\end{equation}
where $\alpha_{j}=\frac{\sum\limits_{k=j}^{N} \tau_+^{k} \tau_-^{N-k}}{\sum\limits_{k=0}^N \tau_+^{k} \tau_-^{N-k}}, \beta_{j}=\frac{\sum\limits_{k=0}^{j-1} \tau_+^{k} \tau_-^{N-k}}{\sum\limits_{k=0}^N \tau_+^{k} \tau_-^{N-k}}$

The training data consist of pointwise unlabeled instances and N-tuples comparisons data, as summarized below:
\begin{itemize}
	\item \textbf{N-tuples Comparisons}: $\mathcal{D}_{j} = \widetilde{\mathcal{D}}_{c_j} =  \{\mathbf{x}_{c_j,i}\}_{i=1}^{n_c} \sim \tilde{p}_j(\mathbf{x})$
	\item \textbf{Pointwise unlabeled data }: $\mathcal{D}_u = \{\mathbf{x}_{u,i}\}_{i=1}^{n_u} \sim p(\mathbf{x})$
\end{itemize}

The empirical risk function  can be reformulated to incorporate $N$-tuple comparison data and pointwise unlabeled data .
\begin{corollary}\label{cor:comp_risk}
The empirical risk for $N_{\text{Comp}}U$ can be rewritten.
\begin{equation}\label{eq.28}
	\begin{split}
		\widehat{R}_{n}(g)
		&= \frac{1}{n_c}\sum_{j=1}^N \sum_{i=1}^{n_c} \left[ \tau_+ C_{1j}\ell(g(\textbf{x}_{j,i}),+1) + \tau_-  C_{2j}\ell(g(\textbf{x}_{j,i}),-1)\right]\\
		&+ \frac{1}{n_u} \sum_{i=1}^{n_u} \left[ \tau_+D_1\ell(g(\textbf{x}_{u,i}),+1) + \tau_-D_2\ell(g(\textbf{x}_{u,i}),-1)\right]\\
		&=\widehat{R}_{cu}(g).
	\end{split}
\end{equation}
with coefficients:
\begin{equation}\label{eq.29}
	\begin{split}
		C_{1j} & = \dfrac{\alpha_{j} \gamma_3 - \beta_{j} \gamma_2}{\gamma_1   \gamma_3 - \gamma_2^2},\quad
		C_{2j}  = \dfrac{-\alpha_{j} \gamma_2 + \beta_{j} \gamma_1}{\gamma_1 \gamma_3 - \gamma_2^2},\\
		D_1 & = \frac{\tau_{+} \gamma_3 - \tau_{-} \gamma_2}{\gamma_1 \gamma_3 - \gamma_2^2},\quad
		D_2  = \frac{-\tau_{+} \gamma_2 + \tau_{-} \gamma_1}{\gamma_1 \gamma_3 - \gamma_2^2}. \\
	\end{split}
\end{equation}
where: $\gamma_1 = \sum\limits_{j=1}^{N}\alpha_{j}^{2}+\tau_+^2, \gamma_2 =\sum\limits_{j=1}^{N}\alpha_{j}\beta_{j}+\tau_+\tau_-, \gamma_3 = \sum\limits_{j=1}^{N}\beta_{j}^{2}+\tau_-^2$.
\end{corollary}

Then, the estimation error bound for $N_{\text{Comp}}U$ learning is given.
\begin{corollary}\label{cor:comp_bound}
 Let
$\widehat{g}_{cu} = argmin_{g\in\mathcal{G}} \widehat{R}_{cu}(g)$ be the $N_{comp}U$ empirical classifier, for any $\delta > 0$, with probability at least $1-\delta$:
	\begin{equation}\label{eq.30}
		\begin{split}
			R(\hat{g}_{cu})- R(g^*) \leq\frac{K_{n}}{\sqrt{n_{c}}}
			+\frac{K_{u}}{\sqrt{n_{u}}},
		\end{split}
	\end{equation}
where $K_n$ and $K_u$ are derived by plugging Eq.~\eqref{eq.29} into the general bound form of Eq.~\eqref{eq.25}.
\end{corollary}

This demonstrates that  the risk \( R(\hat{g}_{cu}) \) converges to the optimal risk \( R(g^*) \) at the rate \( \mathcal{O}\left(\frac{1}{\sqrt{n_{c}}} + \frac{1}{\sqrt{n_u}}\right) \), as \( n_{c}, n_u \to \infty \).

\noindent\textit{Proof Sketch.} 
Corollaries~\ref{cor:comp_risk} and~\ref{cor:comp_bound} are immediate results of Theorems~\ref{thm:asy_general_risk} and~\ref{thm:asy_general_bound},  with coefficients $\alpha_j$ and $\beta_j$ specified in Eq.~~\eqref{eq.4}.

\subsection{Case 2: N-tuples similarities and unlabeled learning}\label{sec5.2}
\textit{N-tuples similarities}\cite{0Learning} refer to a collection of \(N\) instances that all belong to the same class. The associated label space is constrained to:
\begin{equation}\label{eq.6}
	\mathcal{Y}^{\text{sub}} = \mathcal{Y}^{\text{sim}} = \left\{(+1, +1, \dots, +1),\, (-1, -1, \dots, -1)\right\},
\end{equation}

which implies that all instances within a tuple are either positive or negative.

We define the N-tuples similarity dataset as \(\mathcal{D}_{s} = \left\{(\mathbf{x}_{s_{1},i}, \dots, \mathbf{x}_{s_{N},i})\right\}_{i=1}^{n_{s}}\), where each of the \(n_s\) tuples consists of \(N\) instances from the same class, ensuring label consistency. We further flatten the tuples into a pointwise dataset \(\widetilde{\mathcal{D}}_{s} = \left\{\mathbf{x}_{s,i}\right\}_{i=1}^{n_{s}N}\), where each instance is treated independently for downstream learning tasks. $\mathcal{\widetilde{D}}_{s}$ are sampled from:
\begin{equation}\label{eq.7}
	\begin{split}
		\tilde{p}_{s}(\textbf{x})=\frac{\tau_{+}^{N}p_{+}(\textbf{x})+\tau_{-}^{N}p_{-}(\textbf{x})}{\tau_{+}^{N}+\tau_{-}^{N}}.
	\end{split}
\end{equation}

Combined with pointwise unlabeled data , 
\begin{itemize}
	\item \textbf{N-tuples Similarities }: $\mathcal{D}_{j} = \widetilde{\mathcal{D}}_{s} =  \{\mathbf{x}_{s,i}\}_{i=1}^{n_sN} \sim \tilde{p}_s(\mathbf{x})$
	\item \textbf{Pointwise unlabeled data }: $\mathcal{D}_u = \{\mathbf{x}_{u,i}\}_{i=1}^{n_u} \sim p(\mathbf{x})$
\end{itemize}

The corresponding risk estimator and generalization bound under this setting are presented in corollaries~\ref{cor:sim_risk} and~\ref{cor:sim_bound}.
\begin{corollary}[Theorem 5 in \cite{0Learning}]\label{cor:sim_risk}
The rewritten risk function is: 
\begin{equation}\label{eq.9}
	\begin{split}
		&\widehat{R}_{s}(g)=\frac{\tau_{+}^{N}+\tau_{-}^{N}}{(\tau_{+}^{N-1}-\tau_{-}^{N-1})n_{s}^{n}N}\sum_{i=1}^{n_{s}^{n}N}[\ell(g(\textbf{x}),+1)-\ell(g(\textbf{x}),-1)]\\
		&+\frac{1}{n_{u}}\sum_{i=1}^{n_{u}}[-\frac{\tau_{-}^{2}}{2\tau_{+}-1}\ell(g(\textbf{x}),+1)
		+\frac{\tau_{+}^{2}}{2\tau_{+}-1}\ell(g(\textbf{x}),-1)],\\
	\end{split}
\end{equation}
\end{corollary}

\begin{corollary}[Theorem 6 in \cite{0Learning}]\label{cor:sim_bound}
Let
$\widehat{g}_{s} = argmin_{g\in\mathcal{G}} \widehat{R}_{s}(g)$ be the  empirical classifier, for any $\delta > 0$, with probability at least $1-\delta$:
	\begin{equation}\label{label:sim_error}
		\begin{split}
			&R(\hat{g}_{s})-R(g^{\ast})\leq K_{n}\frac{1}{\sqrt{n_{s}^{n}N}}K_{u}\frac{1}{\sqrt{n_{u}}}
		\end{split}
	\end{equation}
	where $K_{n}=\frac{4(\tau_{+}^{N}+\tau_{-}^{N})}{\tau_{+}^{N-1}-\tau_{-}^{N-1}}(2\rho C_{\mathcal{G}}+C_{\ell}\sqrt{\frac{1}{2}\ln\frac{4}{\delta}})$
	and $K_{u}=4\rho C_{\mathcal{G}}+C_{\ell}\sqrt{2\ln\frac{4}{\delta}}$.
\end{corollary}
\subsection{Case 3: Mixed-class N-tuples and unlabeled learning}\label{sec5.3}
In this setting, each N-tuple is known to contain a mixture of positive and negative class instances, but the exact number or positions of the positive samples are not specified. Formally, the constraint is: 
\begin{equation}\label{eq.31}
	\mathcal{Y}^{\text{sub}} = \mathcal{Y}^{\text{mix}} = \{ \mathbf{y} \in \{-1,1\}^N \mid \mathbf{y} \neq \mathbf{-1}, \mathbf{y} \neq \mathbf{1} \}.
\end{equation}

Let $\widetilde{\mathcal{D}}_{m} = \left\{\mathbf{x}_{m,i}\right\}_{i=1}^{n_m N}$ be the pointwise dataset induced from $n_m$ mixed-class $N$-tuples, where each instance is independently drawn from $\tilde{p}_j(\mathbf{x})$.
\begin{theorem} \label{the:mix_distribution}
	Under the mixed-class learning framework, the marginal distribution \(\tilde{p}_j(\mathbf{x})\) in Eq.~\eqref{eq.13} admits the following coefficients:
	\begin{equation}\label{eq.32}
		\alpha = \frac{\sum\limits_{k=1}^{N-1}\binom{N-1}{k}\tau_{+}^{N-k}\tau_{-}^{k}}{\sum\limits_{k=1}^{N-1}\binom{N}{N-k}\tau_{+}^{N-k}\tau_{-}^{k}}, \quad \beta = \frac{\sum\limits_{k=1}^{N-1}\binom{N-1}{N-k}\tau_{+}^{N-k}\tau_{-}^{k}}{\sum\limits_{k=1}^{N-1}\binom{N}{N-k}\tau_{+}^{N-k}\tau_{-}^{k}}.
	\end{equation}
	Under symmetry, \(\alpha_j = \alpha\), \(\beta_j = \beta\) for all \(j\), and matrix \(\mathbf{A} = [\alpha, \beta]\).
\end{theorem}

The detailed proof of the Theorem 6 is provided in the Appendix H.

 The training data consist of: 
\begin{itemize}
	\item \textbf{Mixed-class N-tuples}: $\mathcal{D}_{j}=\widetilde{\mathcal{D}}_{m}\sim \tilde{p}_j(\mathbf{x})$
	\item \textbf{Pointwise unlabeled data }: $\mathcal{D}_u = \{\mathbf{x}_{u,i}\}_{i=1}^{n_u} \sim p(\mathbf{x})$
\end{itemize}

Based on this, the empirical risk for mixed-class and pointwise unlabeled data  is derived as follows.
\begin{corollary}\label{cor:mix_risk}
	 Substituting the coefficients from Eq.~\eqref{eq.32} into Eq.~\eqref{eq.24}, the empirical risk becomes:
	\begin{equation}\label{eq.33}
	\begin{split}
	\widehat{R}_{n}(g)
	&=\frac{\tau_{+}\tau_{-}}{n_{m}N(\alpha\tau_{-}-\beta\tau_{+})}\sum_{i=1}^{n_{m}N}[\mathcal{L}_{\ell}(g(\textbf{x}_{m,i}))]\\
	&+\frac{1}{n_{u}}\sum_{i=1}^{n_{u}}[\mathcal{L}_{u,\ell} g(\textbf{x}_{u,i})]\\
	&=\widehat{R}_{m}(g)
	\end{split}
	\end{equation}
\end{corollary}

The estimation error bound under this risk is given below.
\begin{corollary}\label{cor:mix_bound}	Let
	$\widehat{g}_{m} = argmin_{g\in\mathcal{G}} \widehat{R}_{m}(g)$ be the  empirical classifier, for any $\delta > 0$, with probability at least $1-\delta$:
	\begin{equation}\label{eq.34}
		\begin{split}
			R(\hat{g}_{m})- R(g^*) \leq\frac{K_{n}}{\sqrt{n_{m}N}}
			+\frac{K_{u}}{\sqrt{n_{u}}},
		\end{split}
	\end{equation}
	where $K_n$ and $K_u$  are obtained by plugging the coefficients from Eq.~\eqref{eq.32} into the general bound form of Eq.~\eqref{eq.26}.
\end{corollary}

This result shows that the excess risk \( R(\hat{g}_{m}) \longrightarrow  R(g^*) \) at the optimal rate of \( \mathcal{O}\left(\frac{1}{\sqrt{n_{c}}} + \frac{1}{\sqrt{n_u}}\right) \), as \( n_{m}, n_u \to \infty \).

\noindent\textit{Proof Sketch.} 
Corollaries~\ref{cor:mix_risk} and~\ref{cor:mix_bound} are immediate results of Theorems~\ref{thm:sym_general_risk} and~\ref{thm:sym_general_bound},  with coefficients $\alpha$ and $\beta$ specified in Eq.~~\eqref{eq.32}.  The same applies to  corollaries~\ref{cor:nan_risk} and~\ref{cor:nan_bound}. Thus, the proofs are omitted.
 
\subsection{Case 4: Not-All-Negative N-Tuples and unlabeled learning}\label{sec5.4}
In this setting, each \(N\)-tuple is weakly labeled with the information that at least one instance is from the positive class, while the exact label configuration is unknown. Formally, the weak label constraint is defined as:
\begin{equation}\label{eq.35}
	\mathcal{Y}^{\text{sub}} = \mathcal{Y}^{\text{nan}} = \{ \mathbf{y} \in \{-1,1\}^N \mid \mathbf{y} \neq \mathbf{-1}\}.
\end{equation}
Let $\widetilde{\mathcal{D}}_{e} = \{\mathbf{x}_{e,i}\}_{i=1}^{Nn_e}$ denote the pointwise dataset induced from $n_e$ not-all-negative $N$-tuples, where each instance is independently drawn from the marginal distribution $\tilde{p}_j(\mathbf{x})$.
\begin{theorem}Under the mixed-class weak supervision setting, the marginal distribution \(\tilde{p}_j(\mathbf{x})\)  defined in Eq.~\eqref{eq.13} satisfies:
	\begin{equation}\label{eq.36}
		\alpha= \frac{\tau_{+}^{N} + \sum\limits_{k=1}^{N-1}\binom{N-1}{k}\tau_{+}^{N-k}\tau_{-}^{k}}{1-\tau_{-}^{N}},\quad \beta = \frac{\sum\limits_{k=1}^{N-1}\binom{n-1}{N-k}\tau_{+}^{N-k}\tau_{-}^{k}}{1-\tau_{-}^{N}}.
	\end{equation}
	Under symmetry, \(\alpha_j = \alpha\), \(\beta_j = \beta\) for all \(j\), and the coefficient matrix simplifies to \(\mathbf{A} = [\alpha, \beta]\).
\end{theorem}

A detailed proof of the theorem is presented in the Appendix~I.

 The training data consist of: 
\begin{itemize}
	\item \textbf{Not-all-negative N-tuples}: $\mathcal{D}_{j} =\widetilde{\mathcal{D}}_{e}\sim \tilde{p}_j(\mathbf{x})$
	\item \textbf{Pointwise unlabeled data }: $\mathcal{D}_u = \{\mathbf{x}_{u,i}\}_{i=1}^{n_u} \sim p(\mathbf{x})$
\end{itemize}

Based on these coefficients, the empirical risk can be constructed as follows.
\begin{corollary}\label{cor:nan_risk}
Incorporating the coefficients from Eq.~\eqref{eq.36} into Eq.~\eqref{eq.24}, the risk function can be rewritten as:
	\begin{equation}\label{eq.37}
		\begin{split}
			\widehat{R}_{n}(g)
			&=\frac{\tau_{+}\tau_{-}}{n_{m}^{n}N(\alpha\tau_{-}-\beta\tau_{+})}\sum_{i=1}^{n_{m}^{n}N}[\mathcal{L}_{\ell}(g(\textbf{x}_{m,i}))]\\
			&+\frac{1}{n_{u}}\sum_{i=1}^{n_{u}}[\mathcal{L}_{u,\ell}g(\textbf{x}_{u,i})]\\
			&=\widehat{R}_{e}(g)
		\end{split}
	\end{equation}
\end{corollary}

The corresponding estimation error bound is as follows.
\begin{corollary}\label{cor:nan_bound} 
	Let
	$\widehat{g}_{e} = argmin_{g\in\mathcal{G}} \widehat{R}_{e}(g)$ be the  empirical classifier, for any $\delta > 0$, with probability at least $1-\delta$:
	\begin{equation}\label{eq.38}
		\begin{split}
			R(\hat{g}_{e})- R(g^*) \leq\frac{K_{n}}{\sqrt{n_{e}}}
			+\frac{K_{u}}{\sqrt{n_{u}}},
		\end{split}
	\end{equation}
	where $K_n$ and $K_u$ are specified by substituting the coefficients from Eq.~\eqref{eq.36} into the general bound form of Eq.~\eqref{eq.26}.
\end{corollary}

This demonstrates that the learned risk \( R(\hat{g}_{e}) \) converges to the optimal risk \( R(g^*) \) at the rate  \( \mathcal{O}\left(\frac{1}{\sqrt{n_{e}}} + \frac{1}{\sqrt{n_u}}\right) \), as \( n_{e}, n_u \to \infty \).
\begin{table}[t!]
	\centering
	\caption{Comparison of Key Parameters}
	\label{table:3}
	\begin{adjustbox}{width=0.5\textwidth} 
		\begin{tabular}{@{}l >{\centering\arraybackslash}m{2.5cm}l@{}}
			\toprule
			\textbf{Task Type} & \bm{$\alpha$} & \bm{$\beta$}  \\
			\midrule
			N-tuples comparisons
			& $\frac{\sum\limits_{k=j}^{N} \tau_+^{k} \tau_-^{N-k}}{\sum\limits_{k=0}^{N} \tau_+^{k} \tau_-^{N-k}}$ 
			& $\frac{\sum\limits_{k=0}^{j-1} \tau_+^{k} \tau_-^{N-k}}{\sum\limits_{k=0}^{N} \tau_+^{k} \tau_-^{N-k}}$ \\
			\midrule
			
			Similar N-tuples
			& $\frac{\tau_{+}^{N}}{\tau_{+}^{N}+\tau_{-}^{N}}$ 
			& $\frac{\tau_{-}^{N}}{\tau_{+}^{N}+\tau_{-}^{N}}$ \\
			\midrule
			
			Mixed-class N-tuples
			&$\frac{\sum\limits_{k=1}^{N-1}\binom{N-1}{k}\tau_{+}^{N-k}\tau_{-}^{k}}{\sum\limits_{k=1}^{N-1}\binom{N}{N-k}\tau_{+}^{N-k}\tau_{-}^{k}}$ 
			& $\frac{\sum\limits_{k=1}\limits^{N-1}\binom{N-1}{N-k}\tau_{+}^{N-k}\tau_{-}^{k}}{\sum\limits_{k=1}\limits^{N-1}\binom{N}{N-k}\tau_{+}^{N-k}\tau_{-}^{k}}$ \\
			\midrule
			
			Not-all-negative N-Tuples
			& $\frac{\tau_{+}^{N} + \sum\limits_{k=1}^{N-1}\binom{N-1}{k}\tau_{+}^{N-k}\tau_{-}^{k}}{1-\tau_{-}^{N}}$
			& $\frac{\sum\limits_{k=1}^{N-1}\binom{n-1}{N-k}\tau_{+}^{N-k}\tau_{-}^{k}}{1-\tau_{-}^{N}}$ \\
			\bottomrule
		\end{tabular}
	\end{adjustbox}
\end{table}

\section{Risk correction}
\subsection{General risk formulation}
Eq.~\eqref{eq.22} defines an empirical risk estimator that may involve negative coefficients, potentially leading to negative risk values. Such negative risk is generally regarded as a sign of overfitting. To mitigate overfitting caused by this issue, prior works~~\cite{lu2020mitigating, kiryo2017positive} have proposed  correction functions. Specifically, \cite{lu2020mitigating} enforced the non-negativity of empirical risk by applying a linear correction unit that clips negative values, whereas \cite{kiryo2017positive} argued that negative terms may still contain useful information for training and thus introduced a consistent correction function. The generalized form of the correction function is defined as follows:
\begin{equation}\label{eq.39}
	\begin{split}
		\bar{R}_{n}(g) = f(\widehat{R}_{n}(g)).
	\end{split}
\end{equation}
where, $f(x) = 
\begin{cases}
	x, & x \geq 0, \\
	k|x|, & x < 0.
\end{cases}
\quad \text{and} \quad k > 0.$

In practice, we employ the rectified linear unit (ReLU), $f(z) = \max(0, z)$, and the absolute value function, $f(z) = |z|$, to regularize the negative risk. The effectiveness of applying correction functions to mitigate the impact of negative risk is demonstrated in the experimental results, as illustrated in Fig.~\ref{fig:1}.

\subsection{Consistency guarantee}
Let $\bar{g}_{n} = argmin_{g\in\mathcal{G}} \bar{R}_{n}(g)$.  In this section, we analyze the consistency of the corrected risk function $\bar{R}_n(g)$ and its associated classifier $\bar{g}_n$. First, based on the assumptions on the correction function, we have $\bar{R}_n(g) \geq \widehat{R}_n(g)$. Therefore, we establish the consistency of the corrected risk $\bar{R}(g)$ in Theorem 8.
\begin{theorem}[Risk Consistency of $\bar{R}(g)$]
Let $\tau = \max(\tau_+, \tau_-)$ denote the scaling factor for class-prior weights,  $L_f = \max\{1, k\}$ denotes the Lipschitz constant of the correction function, and $C_w$ be an upper bound on the weighting coefficients $(C_{jk}, D_i)$. Assume there exists $\epsilon > 0$ such that $R_n(g) \geq \epsilon$. Under these conditions, the bias of $\bar{R}(g)$ decays exponentially as $n \rightarrow \infty$.
\begin{equation}\label{eq.40}
	\begin{split}
		\mathbb{E}[\bar{R}(g)]-R(g)\leq \mathcal{O}(1) \cdot \exp\left(-\frac{2\alpha^2}{(Nn_b + n_c)\Delta^2}\right)
	\end{split}
\end{equation}
where $\Delta = \max\left\{ \frac{2N \tau C_w C_\ell}{n_b},\ \frac{2 \tau C_w C_\ell}{n_u} \right\}$, and $\mathcal{O}(1) = (L_f + 1)(N+1) \tau C_w C_\ell$ is a constant factor.

Moreover, with probability at least  $1 -\delta$, the following holds:
\begin{equation}\label{eq.41}
	\begin{split}
		\left|\bar{R}(g) - R(g)\right| &\leq L_\ell\Delta \sqrt{\frac{\ln (2/\delta)}{2n}} \\
		&+ \mathcal{O}(1) \cdot \exp\left(-\frac{2\alpha^2}{(Nn_b + n_c)\Delta^2}\right)
	\end{split}
\end{equation}
\end{theorem}

A detailed proof is presented in the Appendix~J.

Theorem 8 demonstrates the consistency of $\bar{R}(g)$. Leveraging this result, we can further establish the consistency of the associated classifier.

\begin{theorem}[Classifier Consistency of $\bar{g}_n$] Based on the consistency of $\bar{R}(g)$, with probability at least  $1 -\delta$, the following inequality holds:
	\begin{equation}\label{eq.42}
		\begin{split}
			R(\bar{g}_n) - R(g^*) &\leq 2 L_\ell\Delta \sqrt{\frac{\ln (2/\delta)}{2n}} + 2 \mathcal{O}(1) \cdot \exp\left(-\frac{2\alpha^2}{n\Delta^2}\right) \\
			&+ K_{n}\frac{1}{\sqrt{n_b}}+K_{u}\frac{1}{\sqrt{n_{u}}}
		\end{split}
    \end{equation}
\end{theorem}

A detailed proof is presented in the Appendix~K.

Theorem 9 shows that the learned classifier $\bar{g}_n$ is consistent, as its risk converges to that of the optimal classifier $g^*$ as the sample size increases.
\section{Experiment.}\label{sec6}
This section presents empirical evaluations to demonstrate how the proposed generalized framework performs when instantiated for specific learning tasks.

\subsection{Datasets}
\begin{table*}[t!]
	\begin{center}
		\caption{The average classification accuracy and standard error over 5 trials are reported under varying base datasets and category priors in specific experimental settings. The best performance for each method is highlighted in bold.}
		\label{table:4}
		\begin{tabular}{c|c|c|c|c|c|c|c|c|c} 
			\hline
			&&\multicolumn{4}{c|}{Tasks of N-tuples learning}&\multicolumn{4}{c}{Baseline methods}\\
			\hline
			Prior & Dataset & $N_{comp}U$ & NSU & MNU & $N_{pos}U$ & KM & triplet comparison & NT-Comp & triplet sconf \\
			\hline
			\multirow{4}{*}{$\tau_{+}$=0.8} & MNIST & 95.19(0.91) &\textbf{ 95.31(0.22)} & 91.87(2.11) & 83.09(1.74) &68.57(3.14) & 70.67(4.48) & 89.03(0.33)& 92.01(0.83)\\
			&FASHION & \textbf{96.17(0.36)} & 93.66(2.04) & 88.45(1.21) & 85.11(2.46) & 70.22(1.99) &71.85(1.66) & 90.78(1.88)&86.44(1.45)\\
			&SVHN & 74.40(0.99) & \textbf{78.58(0.26)} & 73.89(1.77) &56.22(3.41) & 53.94(4.44) &63.67(4.48)&67.52(0.87)& 73.25(2.63)\\
			&CIFAR-10 & 77.37(0.99) & \textbf{80.07(0.66)} & 78.66(1.27) & 53.72(4.56) & 59.05(5.72) &64.74(3.57)&73.28(1.23)& 74.22(3.51)\\
			\hline
			\multirow{4}{*}{$\tau_{+}$=0.6} & MNIST & \textbf{92.27(0.42)} & 88.08(2.42)) & 83.14(3.33) & 82.55(4.25) & 68.75(2.08) & 71.78(2.28)&90.14(0.76)&90.74(0.53)\\
			&FASHION & 94.34(0.33) & \textbf{90.61(1.44)}  & 88.21(3.11) & 86.45(2.99) &68.25(2.54) &73.44(2.48)& 91.01(1.56)&91.18(0.68)\\
			&SVHN & 60.08(2.73) & \textbf{67.94(1.93)} &54.75(3.81) & 50.55(1.54)&57.42(3.59) &59.33(1.72) & 66.45(2.14)&72.31(4.21)\\
			&CIFAR-10 & 73.46(1.08) & \textbf{78.71(0.29)} & 63.42(3.02) & 52.36(1.83) & 60.77(6.72) & 61.85(2.77)&71.55(1.02)&70.01(3.19)\\
			\hline
			\multirow{4}{*}{$\tau_{+}$=0.2} & MNIST & \textbf{93.72(0.56)}& 92.59(0.41) & 90.80(0.52) & 86.24(1.42)& 64.72(4.66) &71.44(1.66) &91.59(0.77)&91.79(0.96)\\
			&FASHION & 94.45(0.27) & \textbf{93.33(0.79)} & 87.13(0.87) & 94.10(0.20) & 83.07(3.51) &72.08(3.59)&92.88(1.54)&93.01(1.41) \\
			&SVHN & 66.06(2.05) & \textbf{77.56(2.50)} & 71.37(2.75) & 55.36(2.19) &55.98(4.15) &59.41(2.33)&64.27(3.75)& 75.31(4.22)\\
			&CIFAR-10 & 75.40(2.13) & \textbf{82.27(0.82)} & 81.76(1.85) &56.22(1.55) & 58.01(4.88) &59.55(3.74)&73.48(3.11)& 74.21(2.94)\\
			\hline
		\end{tabular}
	\end{center}
\end{table*}
\begin{figure*}[t!]
	\setlength{\belowcaptionskip}{0.01cm} 
	\begin{minipage}{0.24\linewidth}
		\centerline{\includegraphics[width=5cm]{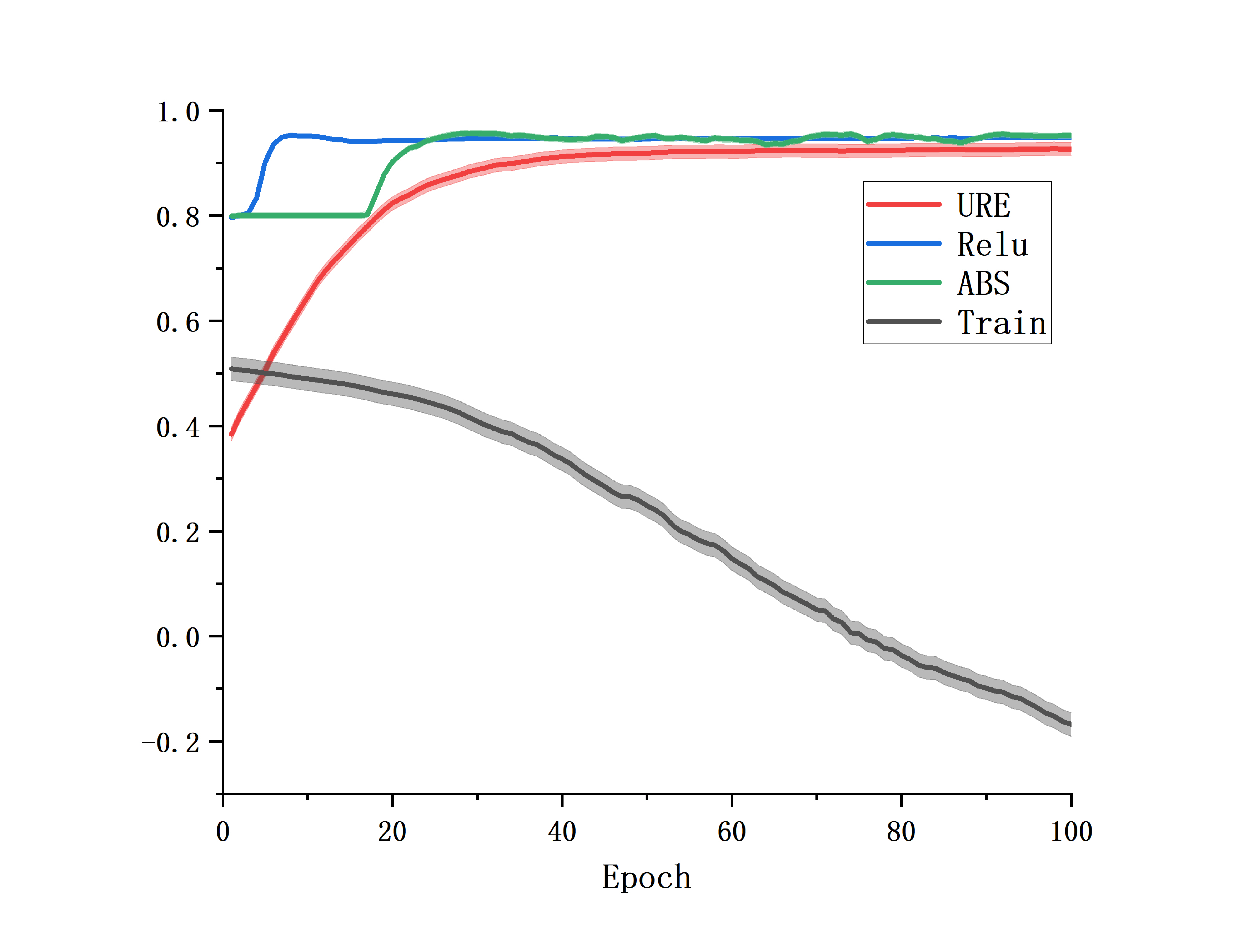}}
		\centerline{(a) MNIST}
	\end{minipage}
	\begin{minipage}{0.24\linewidth}
		\centerline{\includegraphics[width=5cm]{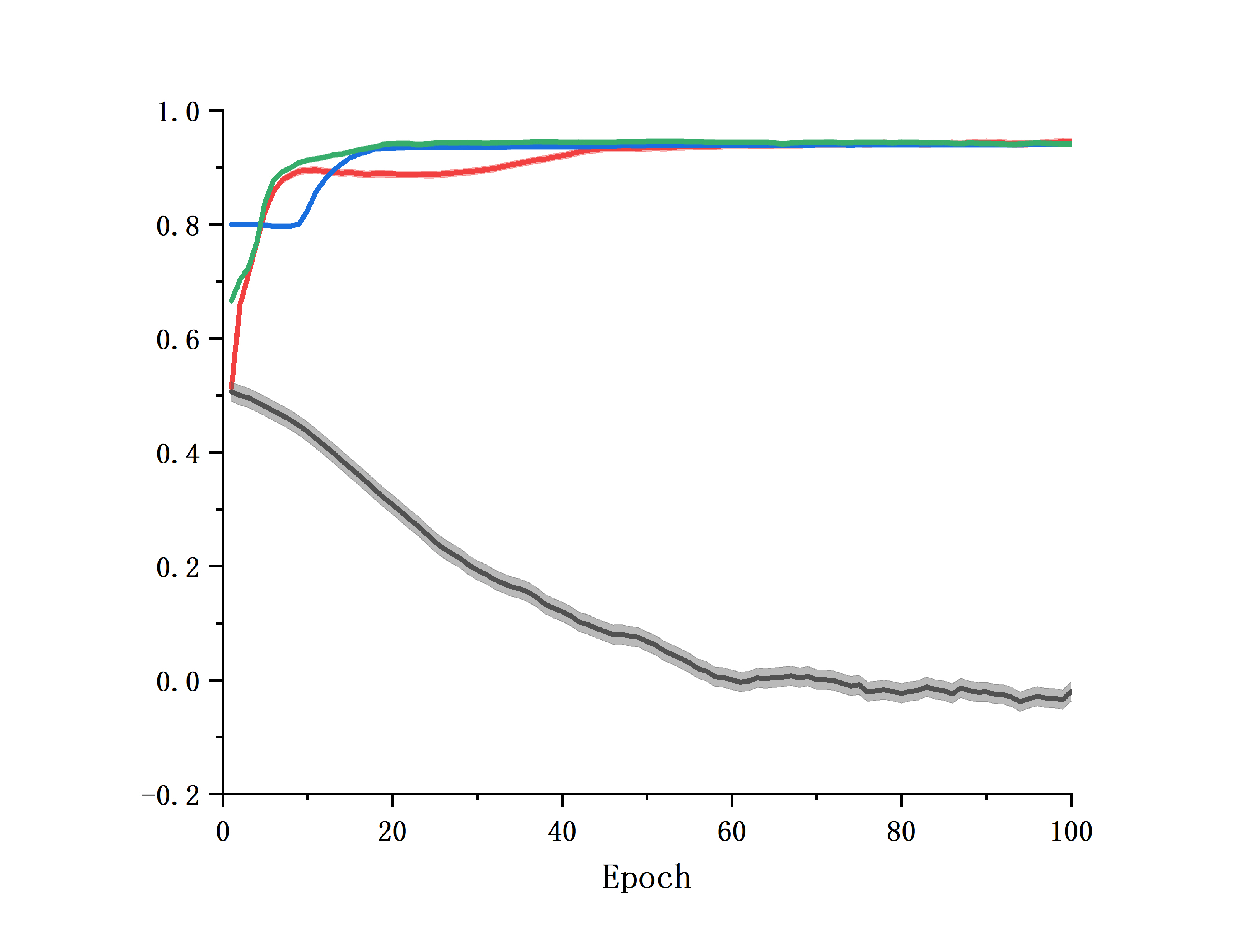}}
		\centerline{(b) Fashion-MNIST}
	\end{minipage}
	\begin{minipage}{0.24\linewidth}
		\centerline{\includegraphics[width=5cm]{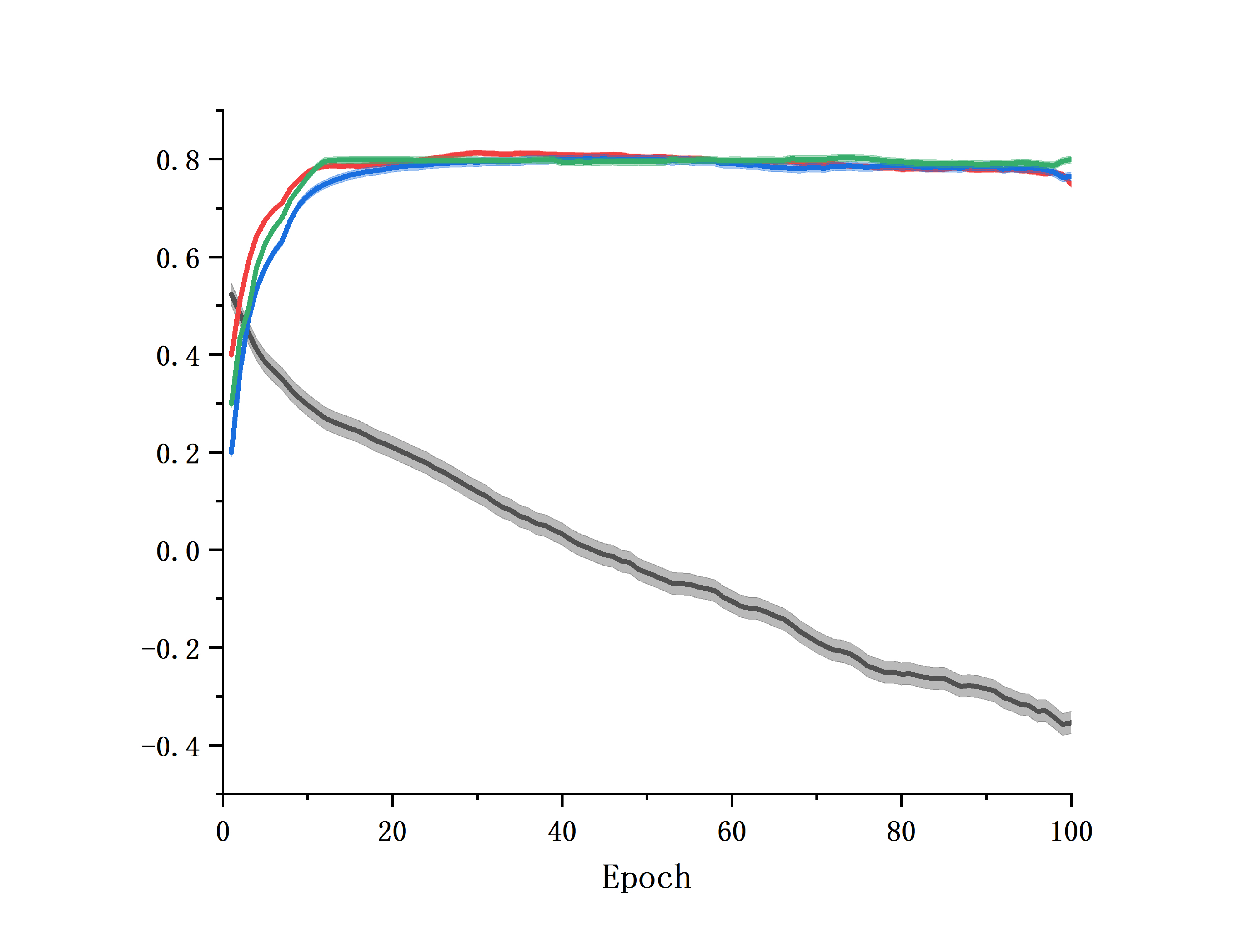}}
		\centerline{(c) SVHN}
	\end{minipage}
	\begin{minipage}{0.24\linewidth}
		\centerline{\includegraphics[width=5cm]{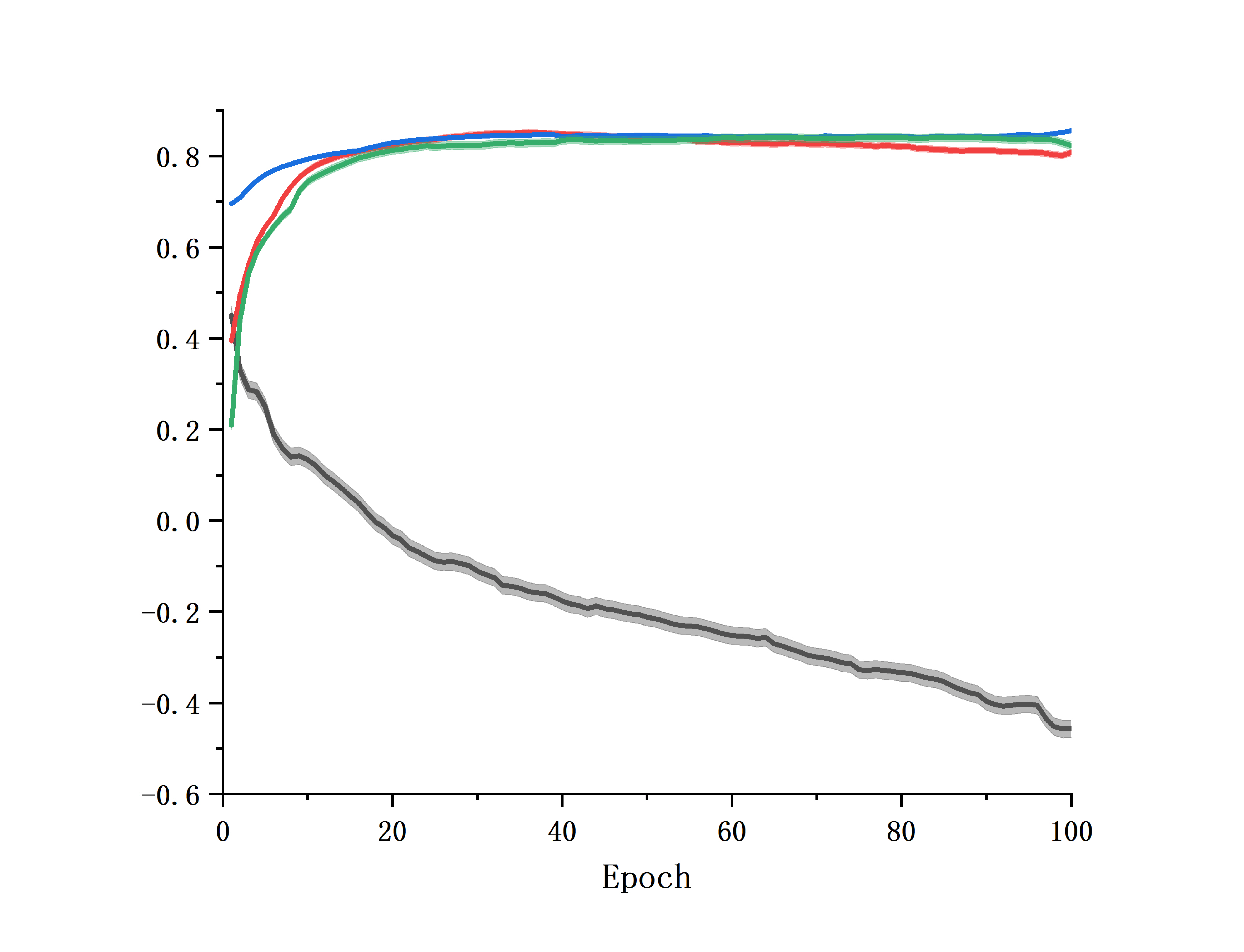}}
		\centerline{(d) Cifar-10}
	\end{minipage}
	\caption{The $N_{\text{comp}}U$ method suffers from overly negative empirical risk on the training set, and the impact of the correction function is accordingly illustrated.}
	\vspace{-0.5em}
	\label{fig:1}
\end{figure*}
\begin{figure*}[t!]
	\setlength{\belowcaptionskip}{0.01cm} 
	\begin{minipage}{0.24\linewidth}
		\centerline{\includegraphics[width=5cm]{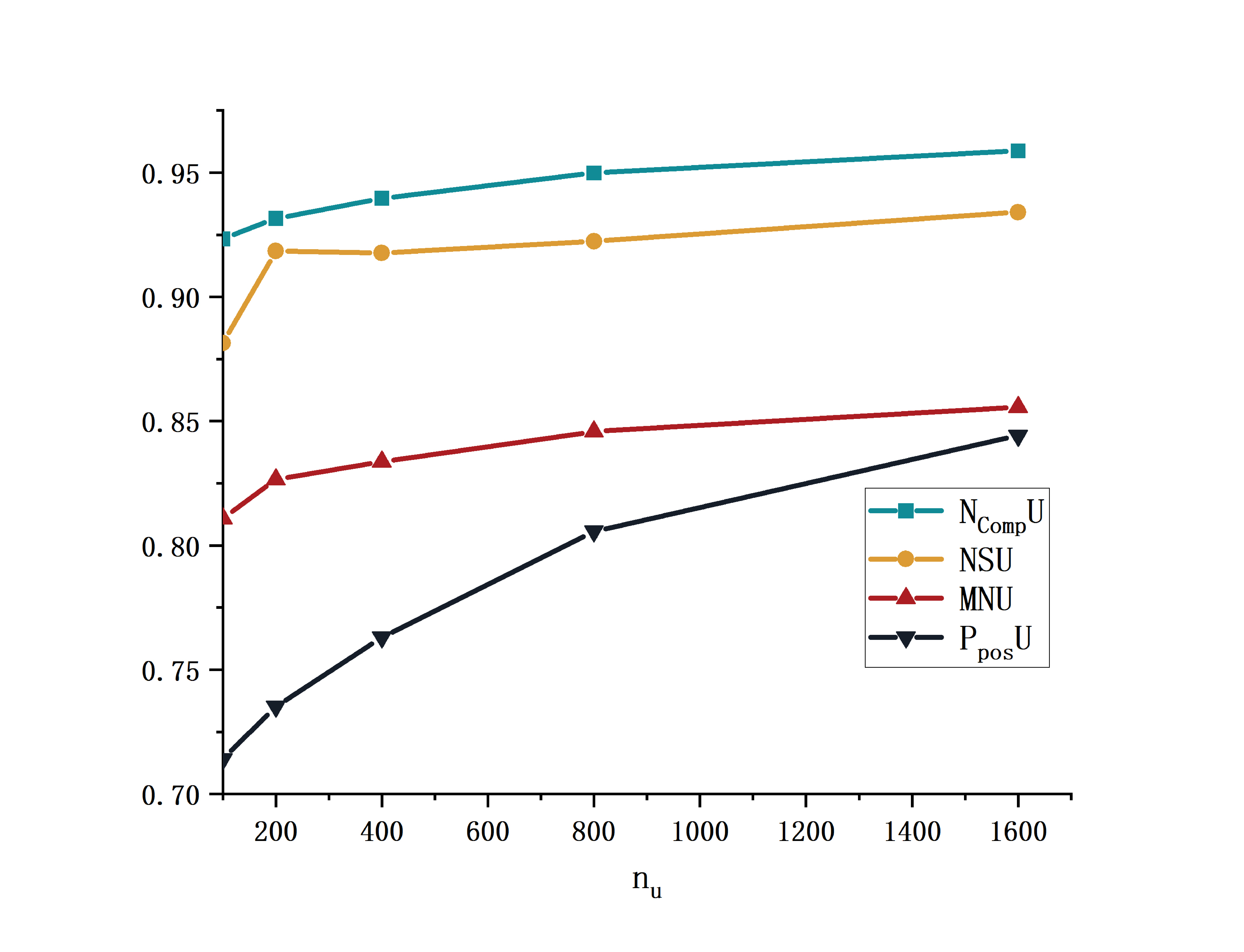}}
		\centerline{(a) MNIST}
	\end{minipage}
	\begin{minipage}{0.24\linewidth}
		\centerline{\includegraphics[width=5cm]{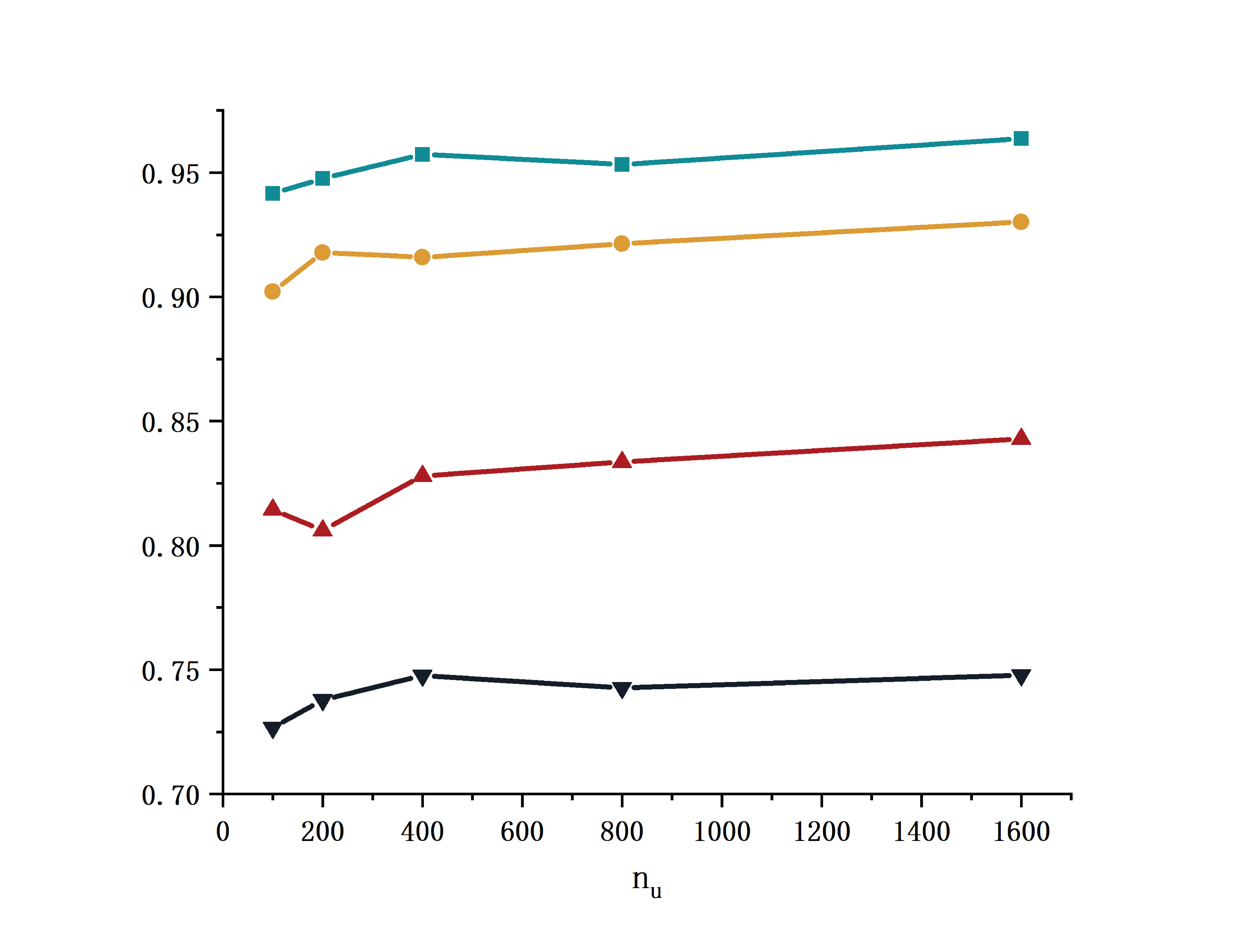}}
		\centerline{(b) Fashion-MNIST}
	\end{minipage}
	\begin{minipage}{0.24\linewidth}
		\centerline{\includegraphics[width=5cm]{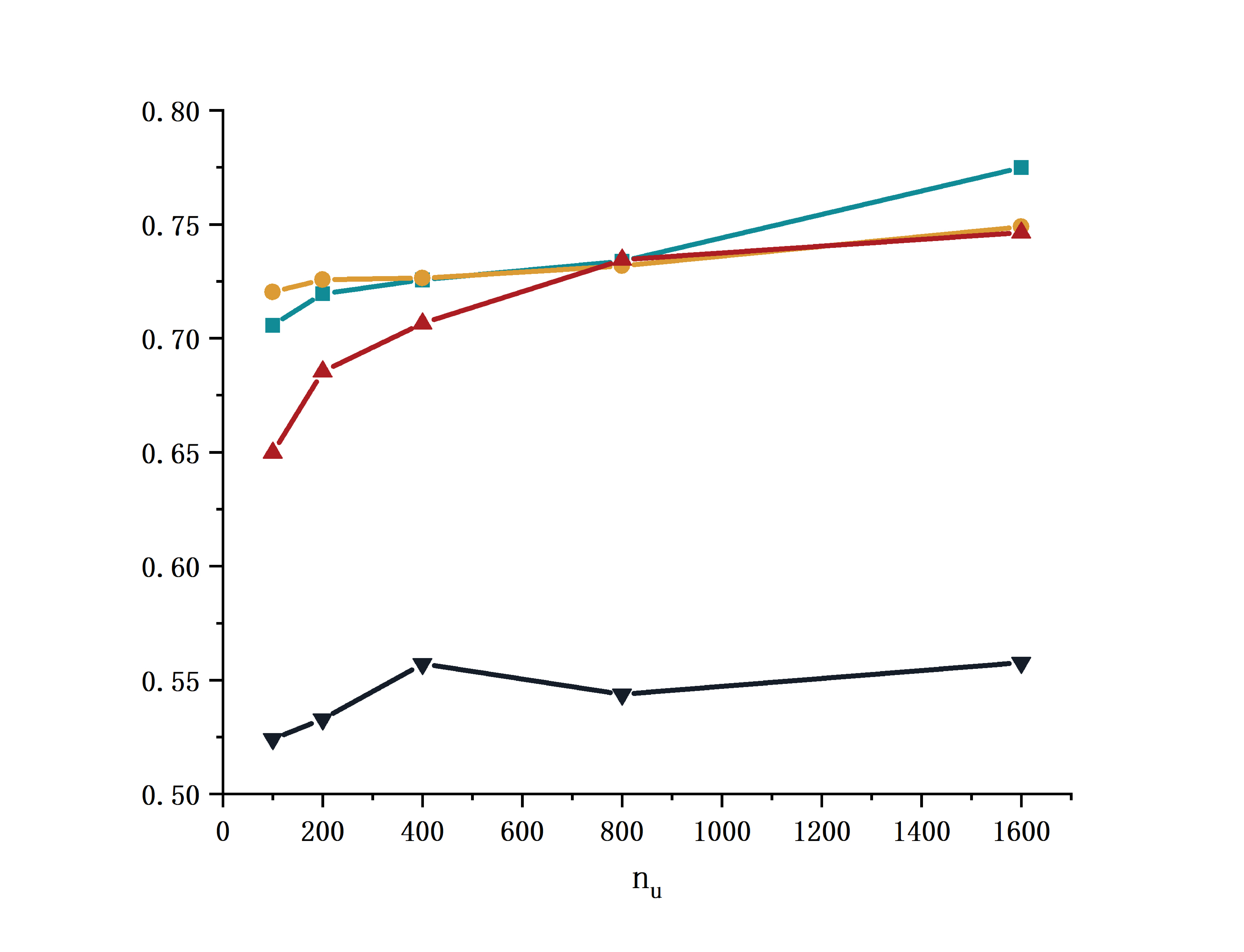}}
		\centerline{(c) SVHN}
	\end{minipage}
	\begin{minipage}{0.24\linewidth}
		\centerline{\includegraphics[width=5cm]{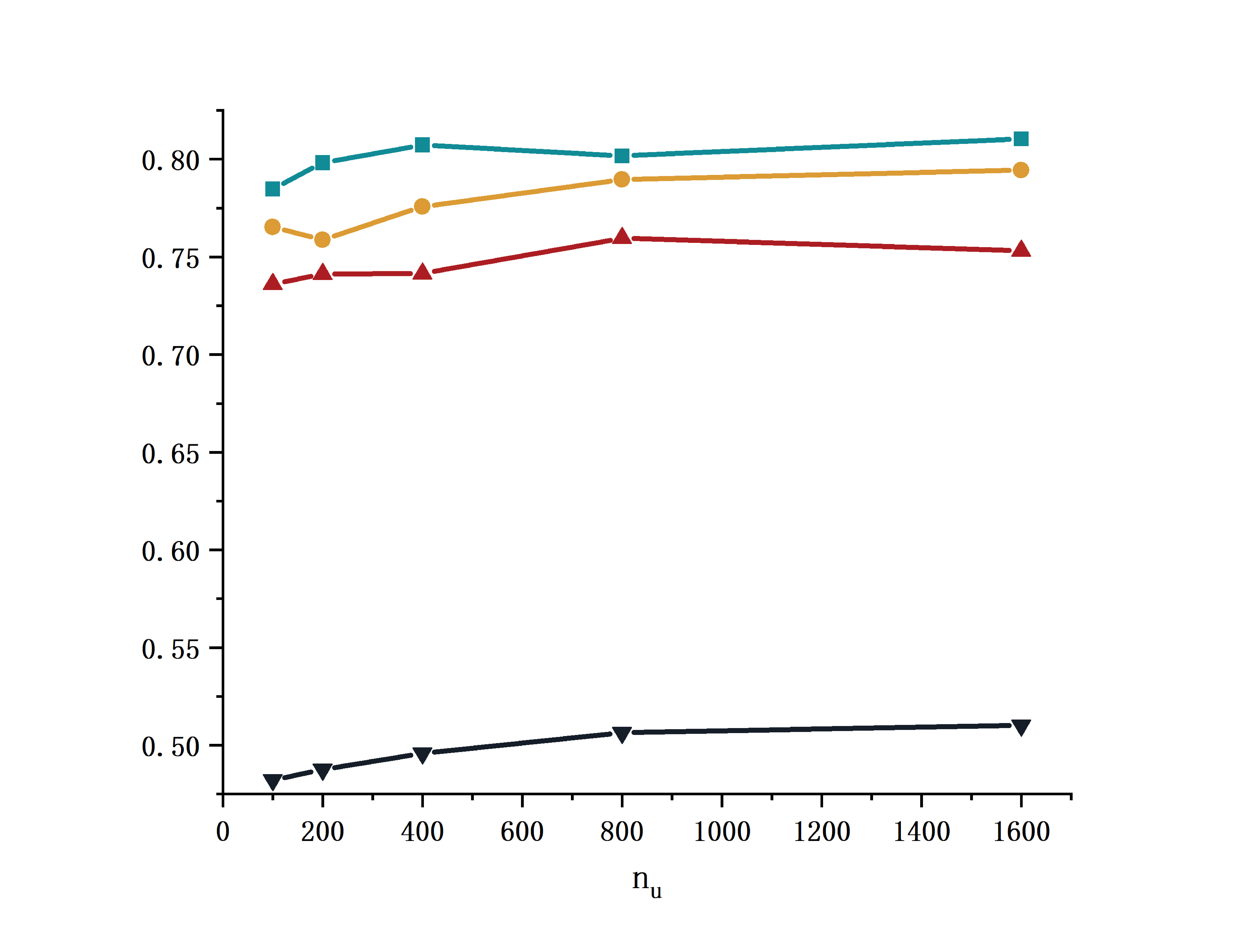}}
		\centerline{(d) Cifar-10}
	\end{minipage}
	\caption{ The impact of varying the number of pointwise unlabeled data  on the performance of different N-tuple learning settings across benchmark datasets. }
	\vspace{-0.5em}
	\label{fig:2}
\end{figure*}
\begin{figure*}[t!]
	\setlength{\belowcaptionskip}{0.01cm} 
	\begin{minipage}{0.24\linewidth}
		\centerline{\includegraphics[width=5cm]{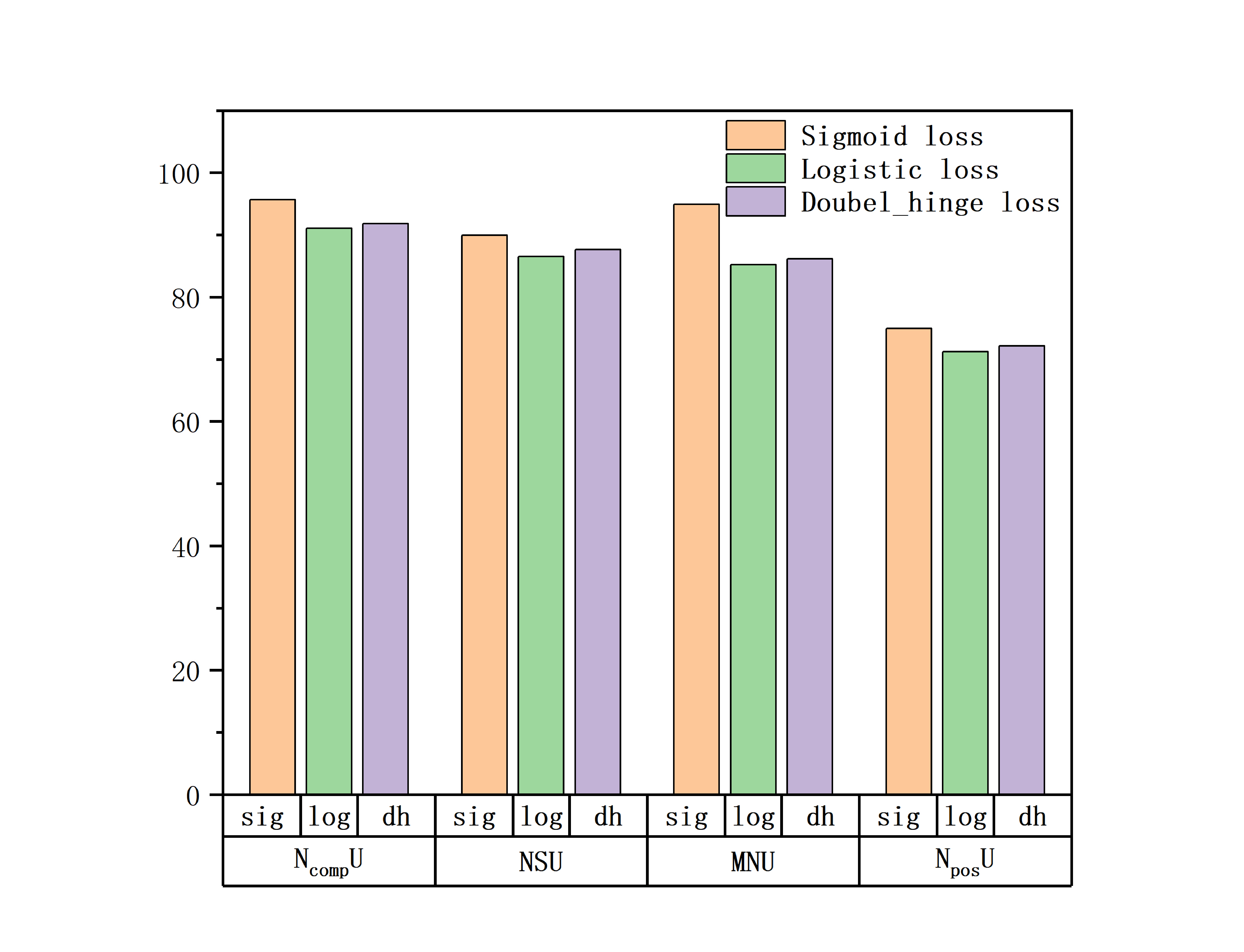}}
		\centerline{(a) MNIST}
	\end{minipage}
	\begin{minipage}{0.24\linewidth}
		\centerline{\includegraphics[width=5cm]{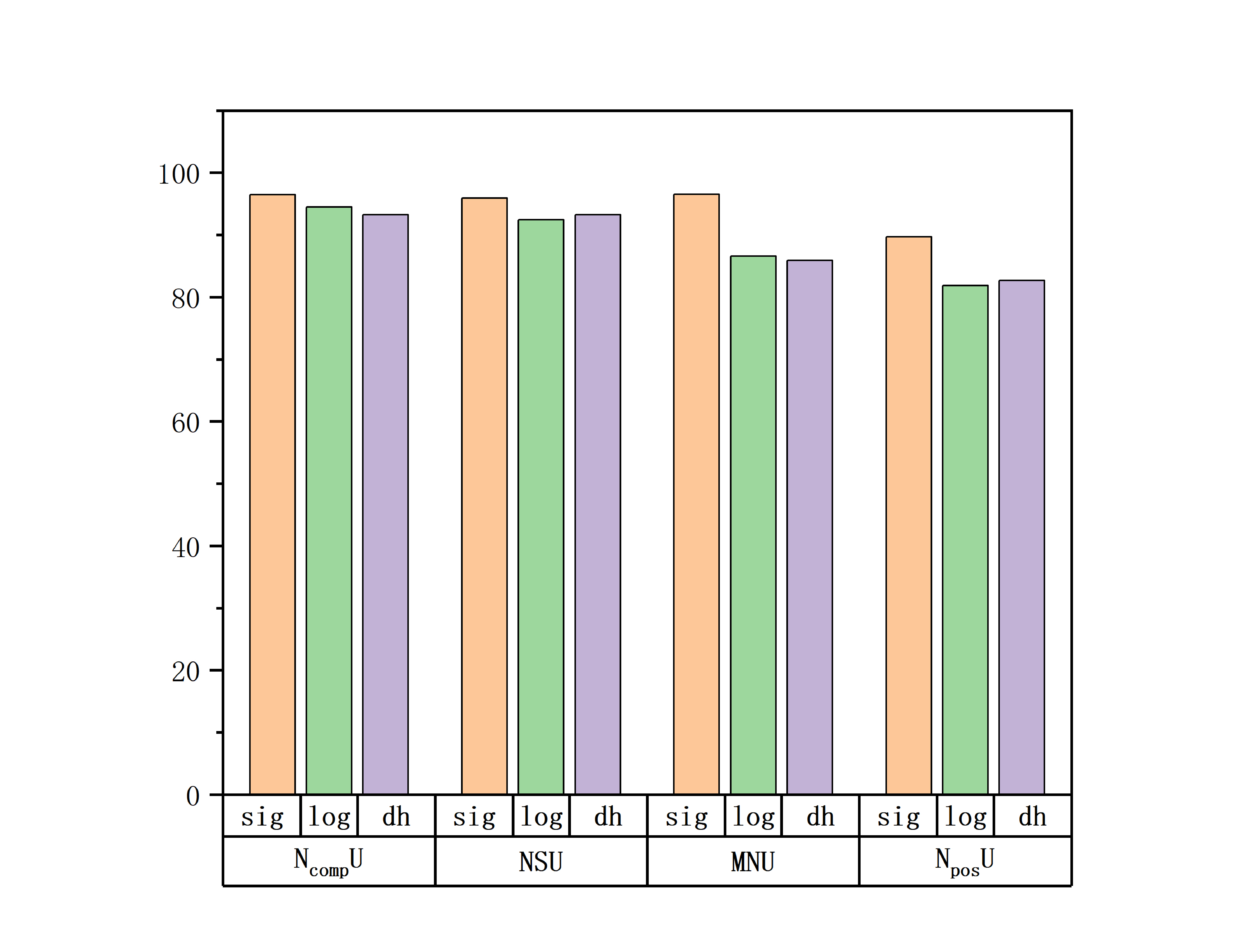}}
		\centerline{(b) Fashion-MNIST}
	\end{minipage}
	\begin{minipage}{0.24\linewidth}
		\centerline{\includegraphics[width=5cm]{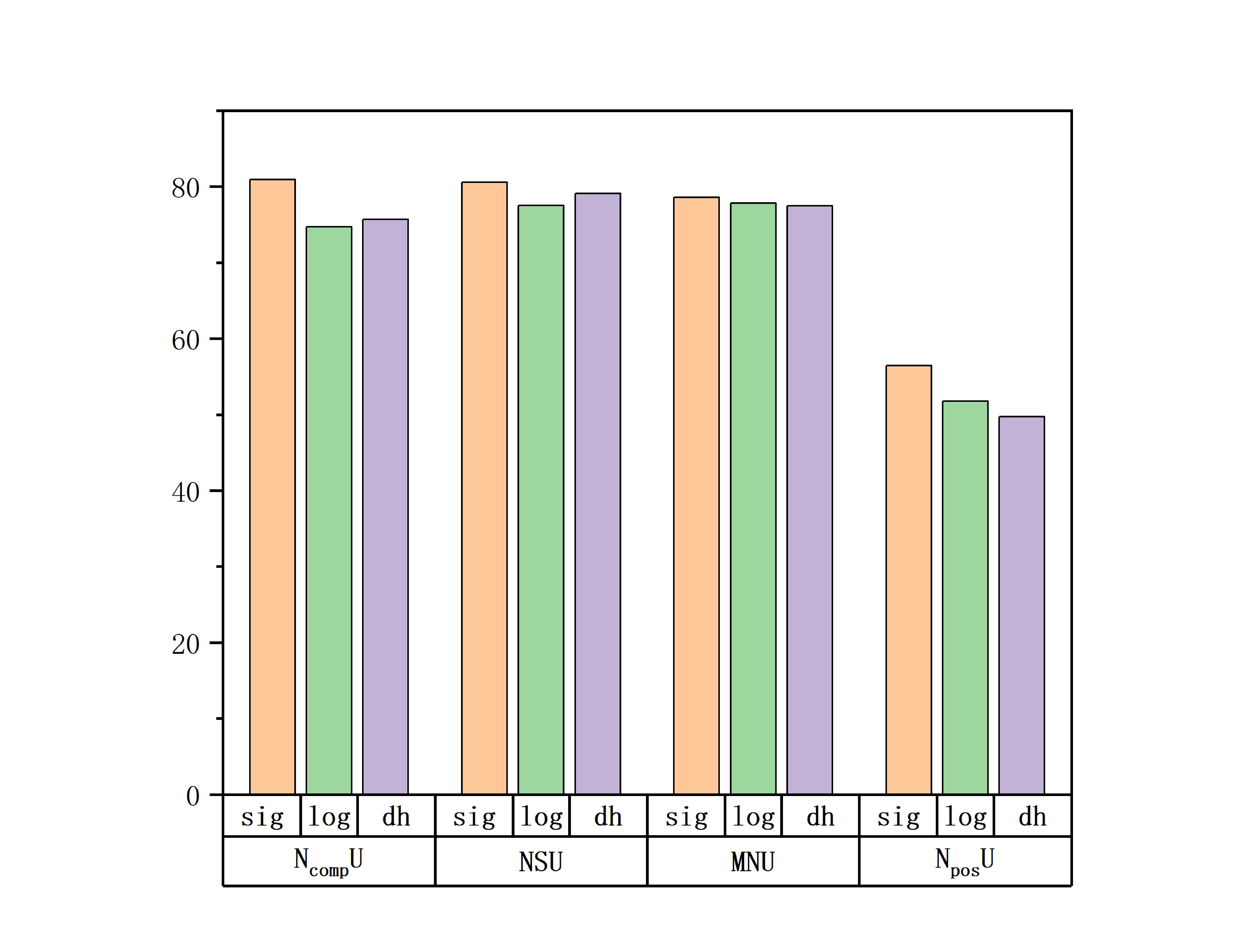}}
		\centerline{(c) SVHN}
	\end{minipage}
	\begin{minipage}{0.24\linewidth}
		\centerline{\includegraphics[width=5cm]{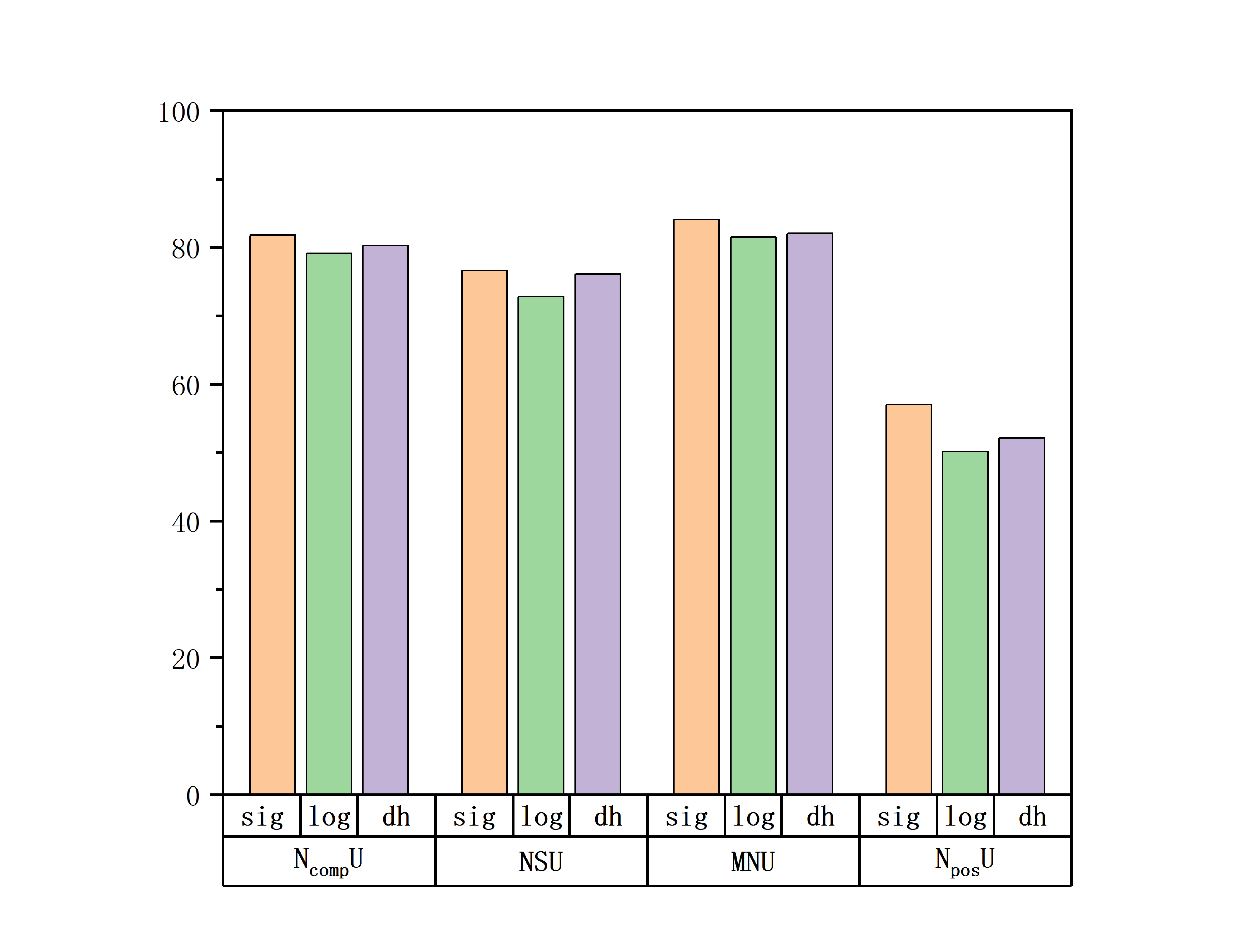}}
		\centerline{(d) Cifar-10}
	\end{minipage}
	\caption{Comparison of classification performance across four N-tuple-based supervision scenarios under various loss functions.}
	\vspace{-0.5em}
	\label{fig:3}
\end{figure*}

In the MNIST dataset, images of even digits are assigned to the positive class, while odd digits are categorized as negative. Each grayscale image has a size of $28\times28$, leading to a flattened input dimension of 784.

For Fashion-MNIST, the categories T-shirt/top, Pullover, Dress, Coat, and Shirt are grouped as the positive class, with the remaining items forming the negative class. The input structure mirrors MNIST, with images of size $28\times28$ and an input dimension of 784.

In SVHN, we follow a similar binary labeling strategy: even digits are considered positive, and odd digits negative. Each color image has dimensions $32\times32\times3$, resulting in an input dimension of 3,072.

For the CIFAR-10 dataset, images depicting airplanes, automobiles, ships, and trucks form the positive class, while the rest are labeled negative. Like SVHN, each image has dimensions $32\times32\times3$, giving an input vector of length 3,072.

\subsection{Baseline methods}
\textbf{NT-Comp\cite{LI2025106894}:} As detailed in Related Works, this baseline operates on N-tuples comparisons data where instances are ranked by their confidence of belonging to the positive class.

\textbf{KM\cite{macqueen1967some}:} K-means is a widely used unsupervised learning method that divides data into $K$ clusters by minimizing the within-cluster sum of squared distances to the centroids. In our setting, we set $K = 2$ to perform binary clustering. The algorithm treats all samples as unlabeled and does not leverage any pairwise similarity or dissimilarity information.

\textbf{Triplet comparison learning\cite{10.1162/neco_a_01262}:} Triplet comparison learning is an emerging paradigm that learn froms comparative feedback data. A typical triplet comparison data, denoted as \((\mathbf{x}_a, \mathbf{x}_b, \mathbf{x}_c)\), conveys the relative similarity information that instance \(\mathbf{x}_a\) is more similar to \(\mathbf{x}_b\) than to \(\mathbf{x}_c\). 

\textbf{M-tuple similarity-confidence learning\cite{li2025binary}:} The proposed Msconf framework extends the Sconf learning paradigm to M-tuples of varying sizes. It leverages similarity-confidence information across multiple instances by jointly modeling their relative confidence levels and inter-instance similarity.

\subsection{The Proposed Methods and Common Setup}
This subsection outlines the implementation details and experimental settings used to evaluate the proposed methods. We assess their performance on several benchmark datasets. For MNIST and Fashion-MNIST, we use a multilayer perceptron (MLP), while for SVHN and CIFAR-10, we adopt a ResNet-based architecture. Training data are sampled from the original datasets and partitioned into positive and negative classes. Subsequently, N-tuples data under different scenarios are constructed according to the class distributions, and combined with unlabeled instances for training.

In our experiments, the loss function $\ell(z)$ is chosen as the sigmoid loss. The performance of the specific tasks is assessed by minimizing the empirical risk in  Eq.~\eqref{eq.28}, ~\eqref{eq.33} and ~\eqref{eq.37}. During training, the learning rate and weight parameters are selected from the range $\{10^{-6}, ..., 10^{-1}\}$. All experiments are implemented in PyTorch and executed on an NVIDIA GeForce RTX 3080 GPU.

\begin{figure}[t!]
	\setlength{\belowcaptionskip}{0.01cm} 
	\centerline{\includegraphics[width=0.5\textwidth]{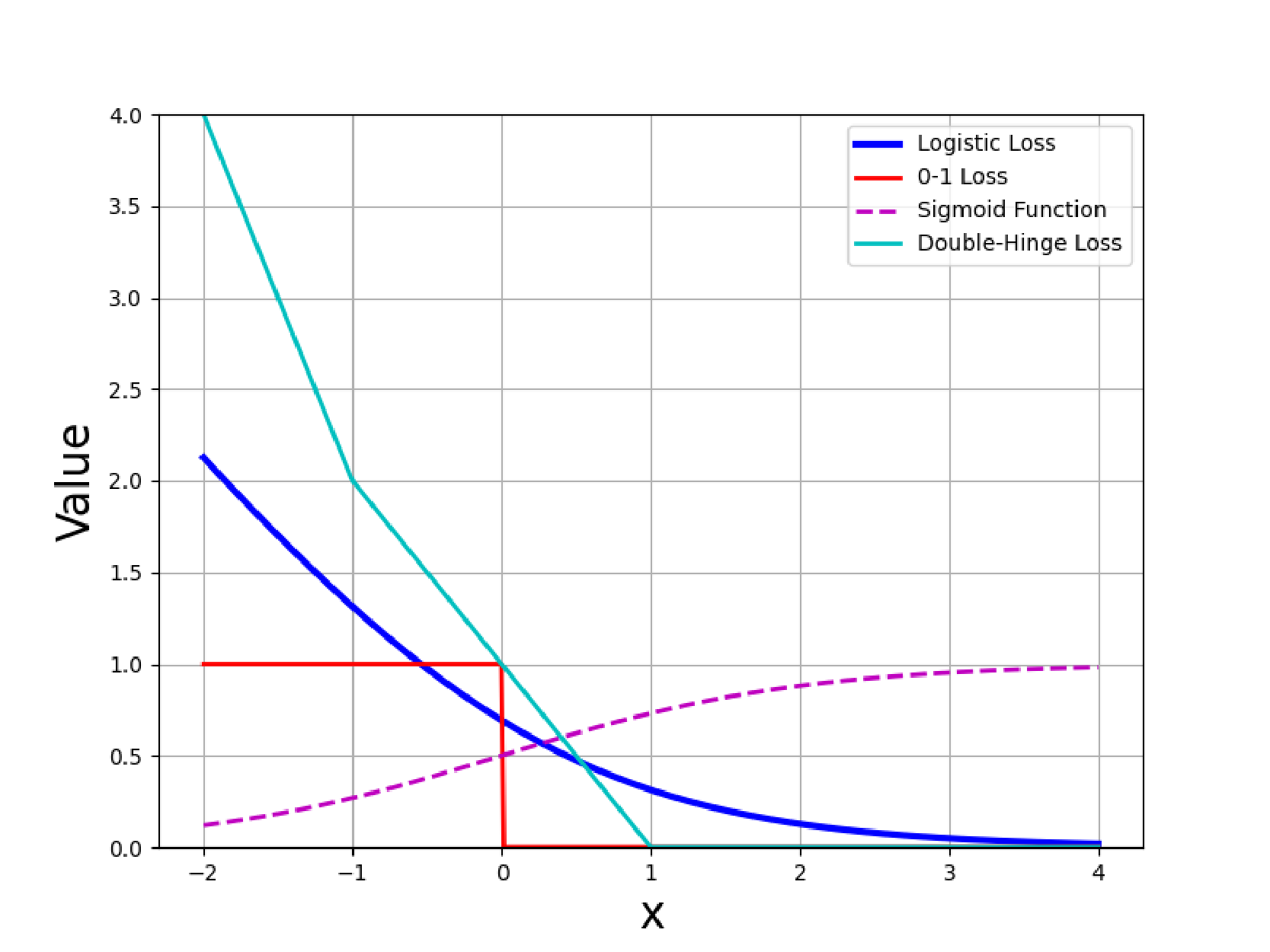}}
	\caption{Comparison of Loss Functions.}
	\label{fig:loss}
\end{figure} 
\subsection{ Experiment Results and Analysis}
We summarize and analyze the performance of our proposed frameworks under different weakly supervised settings using benchmark datasets. The main findings are as follows:

As summarized in Table~\ref{table:4}, iour generalized N-tuple learning framework consistently outperforms baseline methods across all four weak supervision scenarios, highlighting the effectiveness of the proposed risk function in  fully leveraging the weak supervision within N-tuple constraints, as well as the auxiliary signal provided by pointwise unlabeled data. However, Figure~\ref{fig:1} reveals that the empirical training risk may become negative, indicating a risk of overfitting. The improved classification performance after applying our correction function confirms its effectiveness in mitigating such overfitting, thereby emphasizing the necessity of incorporating correction mechanisms in weakly supervised settings.

Figure~\ref{fig:2} demonstrates that increasing the number of pointwise unlabeled samples consistently enhances classification accuracy for all N-tuple-based methods, especially on more challenging datasets. This trend suggests that pointwise unlabeled data play a critical role in improving generalization by facilitating better estimation of class-conditional distributions and decision boundaries. Among the methods evaluated, the $N_{comp}U$ and NSU variants exhibit the strongest performance across all datasets, with accuracy steadily improving as more unlabeled data are introduced. In contrast, the $N_{pos}U$ method yields the lowest performance with limited improvement. This observation indicates that more complex data distributions make it more challenging for the model to capture discriminative features, thereby leading to reduced performance.

Figure~\ref{fig:3} reveals distinct performance patterns across different loss functions, while Figure~\ref{fig:loss} further highlights their respective characteristics. Among the evaluated losses, the sigmoid loss consistently outperforms both logistic and double hinge losses across the four N-tuple weak supervision scenarios. This superior performance may be attributed to its smooth gradient and probabilistic nature, which enable more stable learning under the uncertainty inherent in N-tuple constraints. These results underscore the importance of selecting loss functions that align well with the structural properties of weak supervision frameworks.

Across all methods, we observe a clear performance gap between simpler datasets (e.g., MNIST, Fashion-MNIST) and more complex ones (e.g., SVHN, CIFAR-10), reflecting the inherent limitations of weakly supervised N-tuple learning when applied to high-dimensional and complex data. Nevertheless, the proposed unified framework maintains stable performance trends, demonstrating its adaptability across various levels of data complexity, while also pointing to future opportunities for enhancement in more challenging scenarios.
\bigskip\section{Conclusion.}\label{sec6}
This paper presents a generalized learning framework for N-tuples data aimed at reducing annotation costs in supervised learning. By unifying the generation processes of N-tuples and pointwise unlabeled data  under a common distributional representation, we derive an unbiased empirical risk estimator that subsumes a broad range of existing N-tuples methods. We further instantiate the framework in four representative weakly supervised learning scenarios, illustrating its broad applicability and showing that each can be derived as a specific instance of the proposed general model. The proposed framework not only provides a systematic and theoretically grounded solution for various N-tuples learning scenarios but also demonstrates improved generalization performance through the incorporation of pointwise unlabeled data . This unified perspective offers a practical and versatile approach for handling complex N-tuples structures in real-world applications. Future work will focus on deploying the proposed framework in real-world complex datasets to validate its effectiveness and enhance its scalability in practical applications.

\appendix

\section*{A. Proof of  Lemma 1.}\label{app1}
We derive the sampling distribution $p_{n}(\mathbf{x}_1, \dots, \mathbf{x}_N)$ as follows.
\begin{equation}\label{eq.1}
	\begin{split}
		&p_{n}(\mathbf{x}_1, \dots, \mathbf{x}_N)
		= p\left( (\mathbf{x}_1, \dots, \mathbf{x}_N) \mid (y_1, \dots, y_N) \in  \mathcal{Y}^{\text{sub}} \right) \\
		&= \frac{p\left( (\mathbf{x}_1, \dots, \mathbf{x}_N),\ (y_1, \dots, y_N) \in  \mathcal{Y}^{\text{sub}} \right)}{p\left( (y_1, \dots, y_N) \in  \mathcal{Y}^{\text{sub}} \right)} \\
		&= \frac{\sum\limits_{(y_1, \dots, y_N) \in  \mathcal{Y}^{\text{sub}}} p(\mathbf{x}_1, \dots, \mathbf{x}_N \mid y_1, \dots, y_N) p(y_1, \dots, y_N)}{\sum\limits_{(y_1, \dots, y_N) \in  \mathcal{Y}^{\text{sub}}} p(y_1, \dots, y_N)} \\
		&= \frac{\sum\limits_{(y_1, \dots, y_N) \in  \mathcal{Y}^{\text{sub}}} \left( \prod\limits_{k=1}\limits^N p_{y_k}(\mathbf{x}_k) \cdot \prod\limits_{k=1}\limits^N \tau_{y_k} \right)}{\sum\limits_{(y_1, \dots, y_N) \in \mathcal{Y}^{\text{sub}}} \prod\limits_{k=1}\limits^N \tau_{y_k}} \\
		&= \frac{\sum\limits_{(y_1, \dots, y_N) \in  \mathcal{Y}^{\text{sub}}} \left( \prod\limits_{k=1}\limits^N p_{y_k}(\mathbf{x}_k) \tau_{y_k} \right)}{\sum\limits_{(y_1, \dots, y_N) \in  \mathcal{Y}^{\text{sub}}} \prod\limits_{k=1}\limits^N \tau_{y_k}} 
	\end{split}
\end{equation}
This completes the proof.

\section*{B. Proof of  Theorem 1.}\label{app2}
To derive the marginal distribution of a single instance \(\mathbf{x}_j\) within the group \(\bar{\mathbf{x}}\), we integrate out the other \(N-1\) instances from the joint distribution \(p_n(\bar{\mathbf{x}})\). Specifically, the marginal distribution \(\tilde{p}_j(\mathbf{x}_j)\) is given by:
\begin{align}\label{eq.2}
		\tilde{p}_{j}(\mathbf{x}_j) 
		&= \int p_n(\mathbf{x}_1, \dots, \mathbf{x}_{j-1},\mathbf{x}_{j},\mathbf{x}_{j+1}, \dots, \mathbf{x}_N) \, d\mathbf{x}_{\neq j} \notag\\
		&= \frac{\sum\limits_{\mathbf{y} \in \mathcal{Y}^{\text{sub}}}\left( \int \prod\limits_{k=1}\limits^{N} \tau_{y_k} p_{y_k}(\mathbf{x}_k) d\mathbf{x}_{\neq j}\right) }{\sum\limits_{\mathbf{y} \in \mathcal{Y}^{\text{sub}}} \prod\limits_{k=1}^N \tau_{y_k}} \notag \\
		&= \frac{\sum\limits_{\mathbf{y} \in \mathcal{Y}^{\text{sub}}} \left( p_{y_j}(\mathbf{x}_j)  \prod\limits_{k=1}\limits^{N}\tau_{y_k}\cdot \prod\limits_{k\neq j}\underbrace{\int p_{y_k}(\mathbf{x}_k) d\mathbf{x}_k}_{=1}\right) }{\sum\limits_{\mathbf{y} \in \mathcal{Y}^{\text{sub}}} \prod\limits_{k=1}^N \tau_{y_k}}\notag\\
		&= \frac{\sum\limits_{\mathbf{y} \in \mathcal{Y}^{\text{sub}}} \left(  p_{y_j}(\mathbf{x}_j) \prod\limits_{k=1}\limits^{N} \tau_{y_k}\right) }{\sum\limits_{\mathbf{y} \in \mathcal{Y}^{\text{sub}}} \prod\limits_{k=1}\limits^N \tau_{y_k}}\notag\\
		&= \left( \frac{\sum\limits_{\substack{\mathbf{y} \in \mathcal{Y}^{\text{sub}} \\ y_j = +1}} \prod\limits_{k=1}^N \tau_{y_k}}{\sum\limits_{\mathbf{y} \in\mathcal{Y}^{\text{sub}}} \prod\limits_{k=1}\limits^N \tau_{y_k}} \right) p_{+}(\mathbf{x}_j) \\
		&+ \left( \frac{\sum\limits_{\substack{\mathbf{y} \in \mathcal{Y}^{\text{sub}} \\ y_j = -1}} \prod\limits_{k=1}\limits^N \tau_{y_k}}{\sum\limits_{\mathbf{y} \in\mathcal{Y}^{\text{sub}}} \prod\limits_{k=1}\limits^N \tau_{y_k}} \right) p_{-}(\mathbf{x}_j).
\end{align}

The marginal distribution of an instance $\mathbf{x}_j $ is derived by integrating the joint distribution $\tilde{p}_{j}(\mathbf{x}_j)$ over the other instances, resulting in a weighted mixture of $ p_{+}(\mathbf{x})$ and $ p_{-}(\mathbf{x})$.
\begin{equation}\label{eq.3}
	\tilde{p}_{j}(\mathbf{x}) = 
	\underbrace{
		\left( \frac{\sum\limits_{\substack{\mathbf{y} \in \mathcal{Y}^{\text{sub}} \\ y_j = +1}} \prod\limits_{k=1}^N \tau_{y_k}}{\sum\limits_{\mathbf{y} \in\mathcal{Y}^{\text{sub}}} \prod\limits_{k=1}\limits^N \tau_{y_k}} \right)
	}_{\alpha_j} p_{+}(\mathbf{x})
	+
	\underbrace{
		\left( \frac{\sum\limits_{\substack{\mathbf{y} \in \mathcal{Y}^{\text{sub}} \\ y_j = -1}} \prod\limits_{k=1}\limits^N \tau_{y_k}}{\sum\limits_{\mathbf{y} \in\mathcal{Y}^{\text{sub}}} \prod\limits_{k=1}\limits^N \tau_{y_k}} \right)
	}_{\beta_j} p_{-}(\mathbf{x})
\end{equation}

Thus, Theorem 1 can be proven.

\section*{C. Proof of Lemma 2.}\label{app3}
The linear system relating the observed and latent densities is given by:
\begin{equation}\label{eq.4}
	\mathbf{M} \begin{bmatrix} p_+(\mathbf{x}) \\ p_-(\mathbf{x}) \end{bmatrix} = \begin{bmatrix} \tilde{\mathbf{p}}(\mathbf{x}) \\ p(\mathbf{x}) \end{bmatrix},
\end{equation}
where \(\mathbf{M}\) is an \((N+1) \times 2\) matrix. When \(N \geq 1\), the system is overdetermined.  

To solve for \(p_+(\mathbf{x})\) and \(p_-(\mathbf{x})\), we minimize the squared residual:
\begin{equation}\label{eq.5}
	\left\| \mathbf{M} \begin{bmatrix} p_+(\mathbf{x}) \\ p_-(\mathbf{x}) \end{bmatrix} - \begin{bmatrix} \tilde{\mathbf{p}}(\mathbf{x}) \\ p(\mathbf{x}) \end{bmatrix} \right\|^2.
\end{equation}
The gradient of this quadratic loss with respect to \(\begin{bmatrix} p_+(\mathbf{x}) \\ p_-(\mathbf{x}) \end{bmatrix}\) yields the normal equations:
\begin{equation}\label{eq.6}
	\mathbf{M}^\top \mathbf{M} \begin{bmatrix} p_+(\mathbf{x}) \\ p_-(\mathbf{x}) \end{bmatrix} = \mathbf{M}^\top \begin{bmatrix} \tilde{\mathbf{p}}(\mathbf{x}) \\ p(\mathbf{x}) \end{bmatrix}.
\end{equation}
If \(\mathbf{M}\) has full column rank, \(\mathbf{M}^\top \mathbf{M}\) is invertible, and the solution is uniquely given by:
\begin{equation}\label{eq.7}
	\begin{bmatrix} p_+(\mathbf{x}) \\ p_-(\mathbf{x}) \end{bmatrix} = \left(\mathbf{M}^\top \mathbf{M}\right)^{-1} \mathbf{M}^\top \begin{bmatrix} \tilde{\mathbf{p}}(\mathbf{x}) \\ p(\mathbf{x}) \end{bmatrix}.
\end{equation}

To validate the general solution, we verify its consistency with the symmetric scenario. Here, \(\mathbf{M}\) reduces to:  
\begin{equation}\label{eq.8}
	\mathbf{M} = \begin{bmatrix} \alpha & \beta \\ \tau_+ & \tau_- \end{bmatrix}, \quad \mathbf{M}^\top \mathbf{M} = \begin{bmatrix} \alpha^2 + \tau_+^2 & \alpha\beta + \tau_+\tau_- \\ \alpha\beta + \tau_+\tau_- & \beta^2 + \tau_-^2 \end{bmatrix}.  
\end{equation}  
The determinant of \(\mathbf{M}^\top \mathbf{M}\) is:  
\begin{equation}\label{eq.9}
	\det(\mathbf{M}^\top \mathbf{M}) = (\alpha \tau_- - \beta \tau_+)^2,  
\end{equation}  
which is non-zero if \(\alpha \tau_- \neq \beta \tau_+\) (ensuring invertibility). Substituting into the general solution:  
\begin{equation}\label{eq.10}
	\begin{bmatrix} p_+(\mathbf{x}) \\ p_-(\mathbf{x}) \end{bmatrix} = \frac{1}{\alpha \tau_- - \beta \tau_+} \begin{bmatrix} \tau_- & -\beta \\ -\tau_+ & \alpha \end{bmatrix} \begin{bmatrix} \tilde{p}_j(\mathbf{x}) \\ p(\mathbf{x}) \end{bmatrix},  
\end{equation}  
yields the same result as direct inversion of the \(2 \times 2\) system. This confirms that the general solution specializes correctly to the symmetric case.

This completes the proof. 

\section*{D. Proof of  Theorem 2.}\label{app4}
Substitute the expressions of $p_+(\mathbf{x})$ and $p_-(\mathbf{x})$ into the supervised risk.
\begin{equation}\label{eq.11}
	\begin{split}
		R(g)&= \tau_{+} \mathop{\mathbb{E}}\limits_{p_{+}(\textbf{x})}[\ell(g(\textbf{x}),+1)] + \tau_{-} \mathop{\mathbb{E}}\limits_{p_{-}(\textbf{x})}[\ell(g(\textbf{x}),-1)]\\
		&=\int \left( \sum_{j=1}^N C_{1j} \tilde{p}_j(\mathbf{x}) + D_1 p(\mathbf{x}) \right) \ell(g(\mathbf{x}), +1) d\mathbf{x}\\
		&+ \int \left( \sum_{j=1}^N C_{2j} \tilde{p}_j(\mathbf{x}) + D_2 p(\mathbf{x}) \right) \ell(g(\mathbf{x}), -1) d\mathbf{x}\\
		&=\sum_{j=1}^N \int \left(  C_{1j} \ell(g(\mathbf{x}), +1) + C_{2j} \ell(g(\mathbf{x}), -1) \right)  \tilde{p}_j(\mathbf{x}) d\mathbf{x} \\
		&+\int \left( D_1 \ell(g(\mathbf{x}), +1) + D_2 \ell(g(\mathbf{x}), -1) \right) p(\mathbf{x}) d\mathbf{x} \\
		&=\sum_{j=1}^N \mathop{\mathbb{E}}\limits_{\tilde{p}_j(\textbf{x})}[\tau_{+} C_{1j} \ell(g(\textbf{x}),+1) + \tau_{-} C_{2j} \ell(g(\textbf{x}),-1)]\\
		&+ \mathop{\mathbb{E}}\limits_{p(\textbf{x})} \left[ \tau_{+} D_1 \ell(g(\textbf{x}),+1) + \tau_{-} D_2 \ell(g(\textbf{x}),-1) \right]\\
		&=R_n(g)
	\end{split}
\end{equation}
\section*{E. Proof of  Theorem 3.}\label{app4}
Similarly, the risk function under the symmetry assumption can be reformulated as follows. 
\begin{equation}\label{eq.12}
	\begin{split}
		&R(g)= \tau_{+} \mathop{\mathbb{E}}\limits_{p_{+}(\textbf{x})}[\ell(g(\textbf{x}),+1)] + \tau_{-} \mathop{\mathbb{E}}\limits_{p_{-}(\textbf{x})}[\ell(g(\textbf{x}),-1)]\\
		&=\frac{\tau_{+}}{\alpha \tau_- - \beta \tau_+}\int \left( \tau_- \tilde{p}_j(\mathbf{x}) -  \beta p(\mathbf{x}) \right) \ell(g(\mathbf{x}), +1) d\mathbf{x}\\
		&+ \frac{\tau_{-}}{\alpha \tau_- - \beta \tau_+}\int \left( - \tau_+ \tilde{p}_j(\mathbf{x}) + \alpha p(\mathbf{x}) \right) \ell(g(\mathbf{x}), -1) d\mathbf{x}\\
		&=\frac{\tau_{+}\tau_{+}}{\alpha \tau_- - \beta \tau_+}\int \big( \ell(g(\textbf{x}),+1) - \ell(g(\textbf{x}),-1) \big)\tilde{p}_j(\mathbf{x})d\mathbf{x} \\
		&+\frac{1}{\alpha \tau_- - \beta \tau_+}\int \big( \alpha  \tau_{-}\ell(- \beta \tau_{+}\ell(g(\mathbf{x}, +1 + g(\mathbf{x}, -1))\big) p(\mathbf{x}) d\mathbf{x}\\
		&= \frac{\tau_{+}\tau_{-}}{\alpha \tau_{-} - \beta \tau_{+}} \mathop{\mathbb{E}}\limits_{\mathbf{x} \sim \tilde{p}_{j}(\mathbf{x})} \left[ \ell(g(\textbf{x}),+1) - \ell(g(\textbf{x}),-1) \right] \\
		&+ \frac{1}{\alpha \tau_{-} - \beta \tau_{+}}\mathop{\mathbb{E}}\limits_{\mathbf{x} \sim p(\mathbf{x})} \left[ \alpha  \tau_{-}\ell(g(\mathbf{x}, -1) - \beta \tau_{+}\ell(g(\mathbf{x}, +1) \right]
	\end{split}
\end{equation}

\section*{F. Proof of  Theorem 4.}\label{app5}
Based on the risk function, we define two components of the empirical risk as follows:
\begin{equation}\label{eq.13}
	\begin{split}
		&R_{1}(g)=\sum_{j=1}^N \mathop{\mathbb{E}}\limits_{\tilde{p}_j(\textbf{x})}[\tau_{+} C_{1j} \ell(g(\textbf{x}),+1) + \tau_{-} C_{2j} \ell(g(\textbf{x}),-1)],\\
		&\widehat{R}_{1}(g)\\
		&=\frac{1}{n_b}\sum_{j=1}^N \sum_{i=1}^{n_b} \left[ \tau_{+} C_{1j}\ell(g(\textbf{x}_{j,i}),+1) + \tau_- C_{2j}\ell(g(\textbf{x}_{j,i}),-1)\right],\\
		&R_{2}(g)= \mathop{\mathbb{E}}\limits_{p(\textbf{x})} \left[ \tau_{+} D_1 \ell(g(\textbf{x}),+1) + \tau_{-} D_2 \ell(g(\textbf{x}),-1)\right],\\
		&\widehat{R}_{2}(g)=\frac{1}{n_u} \sum_{i=1}^{n_u} \left[ \tau_+ D_1\ell(g(\textbf{x}_{u,i}),+1) + \tau_-D_2\ell(g(\textbf{x}_{u,i}),-1)\right].
	\end{split}
\end{equation}
We have Lemma 3:

\textbf{Lemma 3:} \textit{The following inequality holds:}
\begin{equation}\label{eq.14}
	\begin{split}
		&R(\hat{g}_{n})-R(g^{\ast})\\
		&\le2\mathop{sup}\limits_{g\in\mathcal{G}}|R_{1}(g)-\widehat{R}_{1}(g)|
		+2\mathop{sup}\limits_{g\in\mathcal{G}}|R_{2}(g)-\widehat{R}_{2}(g)|.\\
	\end{split}
\end{equation}
Proof:
\begin{equation}\label{eq.15}
	\begin{split}
		&R(\hat{g}_{n})-R(g^{\ast})\\
		&=R(\hat{g}_{n})-\widehat{R}_{n}(\hat{g}_{n})+\widehat{R}_{n}(\hat{g}_{n})-\widehat{R}_{n}(g^{\ast})+\widehat{R}_{n}(g^{\ast})-R(g^{\ast})\\
		&=R_{n}(\hat{g}_{n})-\widehat{R}_{n}(\hat{g}_{n})+\widehat{R}_{n}(\hat{g}_{n})-\widehat{R}_{n}(g^{\ast})+\widehat{R}_{n}(g^{\ast})-R_{n}(g^{\ast})\\
		&\leq\mathop{sup}\limits_{g\in\mathcal{G}}|R_{n}(g)-\widehat{R}_{n}(g)|
		+\mathop{sup}\limits_{g\in\mathcal{G}}|\widehat{R}_{n}(g)-\widehat{R}_{n}(g)|\\
		&+\mathop{sup}\limits_{g\in\mathcal{G}}|R_{n}(g)-\widehat{R}_{n}(g)|\\
		&=2\mathop{sup}\limits_{g\in\mathcal{G}}|R_{n}(g)-\widehat{R}_{n}(g)|\\
		&\leq2\mathop{sup}\limits_{g\in\mathcal{G}}|R_{1}(g)-\widehat{R}_{1}(g)|
		+2\mathop{sup}\limits_{g\in\mathcal{G}}|R_{2}(g)-\widehat{R}_{2}(g)|
	\end{split}
\end{equation}

Then, Lemma 4  is crucial to derive an estimating error bound:

\textbf{ Lemma 4:}  \textit{ Let the class function be defined as $\mathcal{G}=\{g:\mathcal{Z}\longrightarrow[0,M]\}$, where
	$(M > 0)$. Then, with probability at least $1-\delta$, the following holds:}
\begin{equation}\label{eq.16}
	\begin{split}
		\mathop{sup}\limits_{g\in\mathcal{G}}|\mathop{\mathbb{E}}[g(\textbf{x})]-\frac{1}{n}\mathop{\sum}\limits_{i=1}^{n}g(\textbf{x}_{i})|
		\le2\Re(\ell\circ\mathcal{G})+\sqrt{\frac{M^{2}\ln\frac{2}{\delta}}{2n}}.
	\end{split}
\end{equation}
where $\{\ell\circ\mathcal{G}|g\in\mathcal{G}\}$ is Rademacher complexity.  By Talagrand lemma:
\begin{equation}\label{eq.17}
	\begin{split}
		\Re(\ell\circ\mathcal{G})\le\rho\Re(\mathcal{G}).
	\end{split}
\end{equation}
Together with $\Re(\mathcal{G})\leq\frac{C_{\mathcal{G}}}{\sqrt{n}}$, we have
\begin{equation}\label{eq.18}
	\begin{split}
		\Re(\ell\circ\mathcal{G})\leq\frac{\rho C_{\mathcal{G}}}{\sqrt{n}}.
	\end{split}
\end{equation}

Based on Lemma 4, the error bounds for the classifier on mixed-class triplets and unlabeled data can be derived from Lemma 5 and 6.

\textbf{ Lemma 5:} \textit{With probability at least $1-\delta$, the following bound holds:}
\begin{equation}\label{eq.19}
	\begin{split}
		&\mathop{sup}\limits_{g\in\mathcal{G}}\mid R_{1}(g)-\widehat{R}_{1}(g)\mid\\
		&\leq \sum_{j=1}^N \left( \tau_{+} C_{1j} + \tau_{-} C_{2j} \right) \left( \frac{2 \rho C_{\mathcal{G}}}{\sqrt{\bar{n}}} + C_{\ell} \sqrt{\frac{\ln(4/\delta)}{2 \bar{n}}} \right)
	\end{split}
\end{equation}

Proof.
\begin{equation}\label{eq.20}
	\begin{split}
		&\mathop{sup}\limits_{g\in\mathcal{G}}|R_{1}(g)-\widehat{R}_{1}(g)|\\
		&= \mathop{sup}\limits_{g\in\mathcal{G}}\big|\sum_{j=1}^N \mathop{\mathbb{E}}\limits_{\tilde{p}_j(\textbf{x})}[\tau_{+} C_{1j} \ell(g(\textbf{x}),+1) + \tau_{-} C_{2j} \ell(g(\textbf{x}),-1)] \\
		&- \frac{1}{n_b}\sum_{j=1}^N \sum_{i=1}^{n_b} \left[ \tau_{+} C_{1j}\ell(g(\textbf{x}_{j,i}),+1) + \tau_- C_{2j}\ell(g(\textbf{x}_{j,i}),-1)\right]\big|\\
		&\leq \sum_{j=1}^N \tau_{+} C_{1j} \sup_{g \in \mathcal{G}} \big| \mathop{\mathbb{E}}_{\tilde{p}_j(\mathbf{x})}[\ell(g(\mathbf{x}),+1)] - \widehat{\mathop{\mathbb{E}}}_{\tilde{p}_j(\mathbf{x})}[\ell(g(\mathbf{x}),+1)] \big| \\
		&+ \sum_{j=1}^N \tau_{-} C_{2j} \sup_{g \in \mathcal{G}} \big| \mathop{\mathbb{E}}_{\tilde{p}_j(\mathbf{x})}[\ell(g(\mathbf{x}),-1)] - \widehat{\mathop{\mathbb{E}}}_{\tilde{p}_j(\mathbf{x})}[\ell(g(\mathbf{x}),-1)] \big|\\
		&\leq \sum_{j=1}^N \left( \tau_{+} C_{1j} + \tau_{-} C_{2j} \right) \left( \frac{2 \rho C_{\mathcal{G}}}{\sqrt{n_b}} + C_{\ell} \sqrt{\frac{\ln(4/\delta)}{2 n_b}} \right)
	\end{split}
\end{equation}

Then, we have Lemma 6:

\textbf{Lemma 6:}\textit{ With probability at least $1-\delta$, the following bound holds:}
\begin{equation}\label{eq.21}
	\begin{split}
		&\mathop{sup}\limits_{g\in\mathcal{G}}|R_{2}(g)-\widehat{R}_{2}(g)|\\
		&\leq \left( \tau_{+} D_1 + \tau_{-} D_2 \right) \left( \frac{2 \rho C_{\mathcal{G}}}{\sqrt{n_u}} + C_{\ell} \sqrt{\frac{\ln(4/\delta)}{2 n_u}} \right),\\
	\end{split}
\end{equation}
The proof of Lemma 6 follows a similar approach to that of Lemma 5.

By combining Lemmas 3, 5, and 6, Theorem 3 can be proven.
\section*{G. Proof of  Theorem 5.}\label{app6}
Analogously, in the case of symmetric data distributions, we begin by defining the following risk functions:
\begin{equation}\label{eq.22}
	\begin{split}
		&R_{s1}(g)=\frac{\tau_{+}\tau_{-}}{\alpha \tau_{-} - \beta \tau_{+}} \mathop{\mathbb{E}}\limits_{\mathbf{x} \sim \tilde{p}_{j}(\textbf{x})} \left[ \mathcal{L}_{\ell}(g(\mathbf{x})) \right], \\
		&\widehat{R}_{s1}(g)=\frac{\tau_{+}\tau_{-}}{N n_b (\alpha \tau_{-} - \beta \tau_{+})} \sum_{i=1}^{N n_b} \mathcal{L}_{\ell}(g(\tilde{\textbf{x}}_{n,i})),\\
		&R_{u2}(g)= \mathop{\mathbb{E}}\limits_{\mathbf{x} \sim p(\mathbf{x})} \left[ \mathcal{L}_{u,\ell}(g(\mathbf{x})) \right],\\
		&\widehat{R}_{u2}(g)=\frac{1}{n_u} \sum_{i=1}^{n_u} \mathcal{L}_{u,\ell}(g(\mathbf{x}_{u,i})).
	\end{split}
\end{equation}

We have,

\textbf{Lemma 7:} \textit{The following inequality holds:}
\begin{equation}\label{eq.23}
	\begin{split}
		&R(\hat{g}_{n})-R(g^{\ast})\\
		&\le2\mathop{sup}\limits_{g\in\mathcal{G}}|R_{s1}(g)-\widehat{R}_{s1}(g)|
		+2\mathop{sup}\limits_{g\in\mathcal{G}}|R_{u2}(g)-\widehat{R}_{u2}(g)|.\\
	\end{split}
\end{equation}
The proof proceeds analogously to that of Lemma 3.

The following inequality is derived as a direct consequence of Lemma 4:

\textbf{ Lemma 8:} \textit{With probability at least $1-\delta$, the following bound holds:}
\begin{equation}\label{eq.24}
	\begin{split}
		&\mathop{sup}\limits_{g\in\mathcal{G}}\mid R_{s1}(g)-\widehat{R}_{s1}(g)\mid
		\leq\frac{\tau_{+}\tau_{-}}{\alpha\tau_{-}-\beta\tau_{+}}(\frac{4\rho C_{\mathcal{G}}}{\sqrt{N n_b}}+2C_{\ell}\sqrt{\frac{\ln\frac{4}{\delta}}{2N\bar{n}}})
	\end{split}
\end{equation}

The following provides an estimation error bound under the setting of unlabeled data.

\textbf{Lemma 9:}\textit{ With probability at least $1-\delta$, the following bound holds:}
\begin{equation}\label{eq.25}
	\begin{split}
		&\mathop{sup}\limits_{g\in\mathcal{G}}|R_{u2}(g)-\widehat{R}_{u2}(g)|\leq \frac{2\rho C_{\mathcal{G}}}{\sqrt{n_{u}}}
		+C_{\ell}\sqrt{\frac{\ln\frac{4}{\delta}}{2n_{u}}},\\
	\end{split}
\end{equation}

Thus, we have,
\begin{equation}\label{eq.26}
	\begin{split}
		R(\hat{g}_{n})-R(g^{\ast})&\leq \frac{2\tau_{+}\tau_{-}}{\alpha\tau_{-}-\beta\tau_{+}}(\frac{4\rho C_{\mathcal{G}}}{\sqrt{N n_b}}\\
		&+2C_{\ell}\sqrt{\frac{\ln\frac{4}{\delta}}{2N\bar{n}}})
		+\frac{4\rho C_{\mathcal{G}}}{\sqrt{n_{u}}}
		+2C_{\ell}\sqrt{\frac{\ln\frac{4}{\delta}}{2n_{u}}}
	\end{split}
\end{equation}
\section*{H. Proof of  Theorem 6.}\label{app6}
The  distribution of mixed-class N-tuples can be shown as:
\begin{equation}\label{eq.27}
	\begin{split}
		&p_{n}(x_{1},x_{2}...x_{N})=p\big((x_{1},x_{2}...x_{N})|(y_{1},y_{2}...y_{N})\in\mathcal{Y}^{\text{mix}}\big)\\
		&=\frac{p\big((x_{1},x_{2}...x_{N}),(y_{1},y_{2}...y_{N})\in\mathcal{Y}^{\text{mix}}\big)}{p\big((y_{1},y_{2}...y_{N})\in\mathcal{Y}^{\text{mix}}\big)}\\
		&=\frac{\sum_{(y_{1},y_{2}...y_{N})\in\mathcal{Y}^{\text{mix}}}p\big(x_{1},x_{2}...x_{N}|(y_{1},y_{2}...y_{N})\big)p(y_{1},y_{2}...y_{N})}{p\big((y_{1},y_{2}...y_{N})\in\mathcal{Y}^{\text{mix}}\big)}\\
		&=\frac{1}{\sum_{i=1}^{N-1}\binom{N}{N-i}\tau_{+}^{N-i}\tau_{-}^{i}}
		\tau_{+}^{N-1}\tau_{-}\\
		&(\prod_{i=1}^{N-1}p_{+}(x_{i})p_{-}(x_{N})+...
		+p_{-}(x_{1})\prod_{i=2}^{N}p_{+}(x_{i}))+...\\
		&+\tau_{+}\tau_{-}^{N-1}(p_{+}(x_{1})\prod_{i=2}^{N}p_{-}(x_{i})+...
		+\prod_{i=1}^{N-1}p_{-}(x_{i})p_{+}(x_{N}).
	\end{split}
\end{equation}

Then, the marginal distribution of $x_{1}$ can be derived  by integrating $x_{2},...,x_{N}$.

\begin{equation}\label{eq.28}
	\begin{split}
		\tilde{p}_{j}(x_{j})&=\frac{1}{\sum_{i=1}^{N-1}\binom{N}{N-j}\tau_{+}^{N-j}\tau_{-}^{j}}\\
		&[\tau_{+}^{N-1}\tau_{-}
		(\int\prod_{j=1}^{N-1}p_{+}(x_{j})p_{-}(x_{N})dx_{2}...dx_{N}
		+...\\
		&+\int p_{-}(x_{1})\prod_{i=2}^{N}p_{+}(x_{j})dx_{2}...dx_{N})+...\\
		&+\tau_{+}\tau_{-}^{N-1}(p_{+}(x_{1})\prod_{i=2}^{N}p_{-}(x_{j})dx_{2}...dx_{N}+...\\
		&+\prod_{j=1}^{N-1}p_{-}(x_{j})p_{+}(x_{N})dx_{2}...dx_{N})]\\
		&=\frac{1}{\sum_{j=1}^{N-1}\binom{N}{N-j}\tau_{+}^{N-j}\tau_{-}^{j}}\Big[\sum_{j=1}^{N-1}\binom{N-1}{j}\tau_{+}^{N-j}\tau_{-}^{j}p_{+}(x_{1})\\
		&+\sum_{j=1}^{N-1}\binom{N-1}{N-j}\tau_{+}^{N-j}\tau_{-}^{j}p_{-}(x_{1})\Big]
	\end{split}
\end{equation}

Thus,
\begin{equation}\label{eq.29}
	\begin{split}
		\tilde{p}_{j}(x)&=\frac{1}{\sum_{i=1}^{N-1}\binom{N}{N-i}\tau_{+}^{N-i}\tau_{-}^{i}}\Big[\sum_{i=1}^{N-1}\binom{N-1}{i}\tau_{+}^{N-i}\tau_{-}^{i}p_{+}(x)\\
		&+\sum_{i=1}^{N-1}\binom{N-1}{N-i}\tau_{+}^{N-i}\tau_{-}^{i}p_{-}(x)\Big]
	\end{split}
\end{equation}

The process of deriving the edit distribution for $x_{2},...,x_{N}$ follows the same approach as for $x_{1}$.
\section*{I. Proof of  Theorem 7.}\label{app7}
The joint distribution of N-tuples containing at least one positive instance is presented as follows:
\begin{equation}\label{eq.30}
	\begin{split}
		&p_{n}(\textbf{x}_{1},\textbf{x}_{2}...\textbf{x}_{N})=p((\textbf{x}_{1},\textbf{x}_{2}...\textbf{x}_{N})|(y_{1},y_{2}...y_{N})\in\mathcal{Y}^{\text{nan}})\\
		&=\frac{p((\textbf{x}_{1},\textbf{x}_{2}...\textbf{x}_{N}),(y_{1},y_{2}...y_{N})\in\mathcal{Y}^{\text{nan}})}{p((y_{1},y_{2}...y_{n})\in\mathcal{Y}^{\text{nan}})}\\
		&=\frac{\sum_{(y_{1},y_{2}...y_{N})\in\mathcal{Y}^{\text{nan}}}p(\textbf{x}_{1},\textbf{x}_{2}...\textbf{x}_{N}|(y_{1},y_{2}...y_{N}))p(y_{1},y_{2}...y_{N})}{p((y_{1},y_{2}...y_{N})\in\mathcal{Y}^{\text{nan}})}\\
		&=\frac{1}{\tau_{+}^{N}+\binom{n}{1}\tau_{+}^{N-1}\tau_{-}...
			+\binom{N}{N-1}\tau_{+}\tau_{-}^{N-1}}
		(\tau_{+}^{N}\prod_{i=1}^{N}p_{+}(\textbf{x}_{i})\\
		&+\tau_{+}^{N-1}\tau_{-}(\prod_{i=1}^{N-1}p_{+}(\textbf{x}_{i})p_{-}(\textbf{x}_{N})+...+p_{-}(\textbf{x}_{1})\prod_{i=2}^{N}p_{+}(\textbf{x}_{i}))
		+...\\
		&+\tau_{+}\tau_{-}^{N-1}(p_{+}(\textbf{x}_{1})\prod_{i=2}^{N}p_{-}(\textbf{x}_{i})+...+\prod_{i=1}^{N-1}p_{-}(\textbf{x}_{i})p_{+}(\textbf{x}_{N})).
	\end{split}
\end{equation}

Then, the marginal distribution of each example can be obtained by performing  integration. For instance, the marginal distribution of example $\textbf{x}_1$, denoted as  $\tilde{p}_m^n(\mathbf{\textbf{x}}_1)$ , can be derived by integrating over $\textbf{x}_2, \dots, \textbf{x}_n$.  
\begin{equation}\label{eq.31}
	\begin{aligned}
		\tilde{p}_{j}(\textbf{x}_{1})&=\frac{\tau_{+}^{N}}{1-\tau_{-}^{N}}\int \prod_{i=1}^{N}p_{+}(\textbf{x}_{i})d\textbf{x}_{2}d\textbf{x}_{3}...d\textbf{x}_{N}\\
		&+\frac{\tau_{+}^{n-1}\tau_{-}}{1-\tau_{-}^{N}}[\int\prod_{i=1}^{n-1}p_{+}(\textbf{x}_{i})p_{-}(\textbf{x}_{N})d\textbf{x}_{2}d\textbf{x}_{3}...d\textbf{x}_{N}\\
		&+...+\int p_{-}(\textbf{x}_{1})\prod_{i=2}^{N}p_{+}(\textbf{x}_{i})d\textbf{x}_{2}d\textbf{x}_{3}...d\textbf{x}_{N}]
		+...+\\
		&+\frac{\tau_{+}\tau_{-}^{N-1}}{1-\tau_{-}^{N}}[\int p_{+}(\textbf{x}_{1})\prod_{i=2}^{n}p_{-}(\textbf{x}_{i})d\textbf{x}_{2}d\textbf{x}_{3}...d\textbf{x}_{N}
		+...\\
		&+\int\prod_{i=1}^{N-1}p_{-}(\textbf{x}_{i})p_{+}(\textbf{x}_{N})d\textbf{x}_{2}d\textbf{x}_{3}...d\textbf{x}_{N}]\\
		&=\frac{1}{1-\tau_{-}^{N}}\Big[\tau_{+}^{N}p_{+}(\textbf{x}_{1})\\
		&+\sum_{i=1}^{N-1}\tau_{+}^{N-i}\tau_{-}^{i}[\binom{N-1}{i}p_{+}(\textbf{x}_{1})+\binom{N-1}{N-i}p_{-}(\textbf{x}_{1})]\Big]\\
		&=	\frac{\tau_{+}^{n}+\sum_{i=1}^{n-1}\tau_{+}^{n-i}\tau_{-}^{i}\binom{n-1}{i}}{1-\tau_{-}^{N}}p_{+}(\textbf{x}_{1})\\
		&+\frac{\sum_{i=1}^{N-1}\tau_{+}^{N-i}\tau_{-}^{i}\binom{N-1}{N-i}}{1-\tau_{-}^{N}}p_{-}(\textbf{x}_{1}).
	\end{aligned}
\end{equation}

Thus,
\begin{equation}\label{eq.32}
	\begin{aligned}
		\tilde{p}_{j}(\textbf{x}_{1})&=\frac{\tau_{+}^{N}+\sum_{i=1}^{N-1}\tau_{+}^{N-i}\tau_{-}^{i}\binom{N-1}{i}}{1-\tau_{-}^{N}}p_{+}(\textbf{x}_{1})\\
		&+\frac{\sum_{i=1}^{n-1}\tau_{+}^{N-i}\tau_{-}^{i}\binom{N-1}{N-i}}{1-\tau_{-}^{N}}p_{-}(\textbf{x}_{1})
	\end{aligned}
\end{equation}

Since the distribution of $\textbf{x}_1, \dots, \textbf{x}_N$ is symmetric,
\begin{equation}\label{eq.33}
	\begin{aligned}
		\tilde{p}_{j}(\textbf{x})&=\frac{\tau_{+}^{N}+\sum_{i=1}^{N-1}\tau_{+}^{N-i}\tau_{-}^{i}\binom{N-1}{i}}{1-\tau_{-}^{N}}p_{+}(\textbf{x})\\
		&+\frac{\sum_{i=1}^{n-1}\tau_{+}^{n-i}\tau_{-}^{i}\binom{N-1}{N-i}}{1-\tau_{-}^{N}}p_{-}(\textbf{x})
	\end{aligned}
\end{equation}

\section*{J. Proof of  Theorem 8.}\label{app8}
\textbf{Definition 2 }
Define the dataset $S_n = \mathcal{D}_n \cup \mathcal{D}_u $.
Given a classifier $g$, we define the  $\Omega_-(g)$ as the set of all datasets $S_n$ for which the empirical risk underestimates the true risk:
$$
\Omega_-(g) \triangleq \left\{ S_n \mid \widehat{R}_n(g) < 0 \right\}
$$
Similarly, we define the  $\Omega_+(g) \triangleq \left\{ S_n \mid \widehat{R}_n(g) > 0 \right\}$.

Based on the above definitions and assumptions, we first present the following lemma.

\textbf{Lemma 10:} \label{lemma:10}
\textit{Assume that there is $\epsilon> 0$  such that $R_n(g) \geq \epsilon$. By assumptions in Theorem~3, the probability measure of $\mathfrak{D}_-(g)$ can be upper bounded by:}
\begin{equation}\label{eq.34}
	\begin{split}
		\mathbb{P}(\Omega_-(g)) \leq \exp\left( -\frac{2\alpha^2}{(Nn_b + n_c) \Delta^2} \right)
	\end{split}
\end{equation}

\textit{Proof:} According to the data generation process:
\begin{equation}\label{eq.35}
	\begin{split}
		p(S_n) = \left( \prod_{j=1}^{n_b} p_{j}(\{(x_{j,k})\}_{k=1}^N) \right) \times \left( \prod_{i=1}^{n_u} p(x_{u,i}) \right)
	\end{split}
\end{equation}

The change in $\widehat{R}_n(g)$ is at most $\frac{2N \tau C_w C_\ell}{n_b}$ when replacing an N-tuple, and at most $\frac{2 \tau C_w C_\ell}{n_u}$ when replacing an unlabeled instance.
Let $\Delta = \max\left\{ \frac{2N \tau C_w C_\ell}{n_b},\ \frac{2 \tau C_w C_\ell}{n_u} \right\}$ denote the maximum change in the empirical risk $\widehat{R}_n(g)$ caused by replacing either an $N$-tuple or an unlabeled instance. According to McDiarmid?s inequality, since modifying any single input (either an $N$-tuple or an unlabeled sample) changes $\widehat{R}_n(g)$ by at most $\Delta$, we have the following concentration bound:
\begin{equation}\label{eq.36}
	\begin{split}
		\mathbb{P}\left( \widehat{R}_n(g) - \mathbb{E}[\widehat{R}_n(g)] \leq -\epsilon \right) \leq \exp\left( -\frac{2\epsilon^2}{(Nn_b + n_c) \Delta^2} \right).
	\end{split}
\end{equation}

Given that the expected risk satisfies $\mathbb{E}[\widehat{R}_n(g)] = R_n(g) \geq \epsilon > 0$, it follows that the probability of the empirical risk underestimating the true risk is bounded by
\begin{equation}\label{eq.37}
	\begin{split}
		\mathbb{P}\left( \Omega_-(g) \right) \leq \exp\left( -\frac{2\epsilon^2}{(Nn_b + n_c)\Delta^2} \right),
	\end{split}
\end{equation}
This completes the proof.

We now proceed to present the proof of Theorem 7.

\textit{Proof:} Based on the definition of a consistent correction function, we have:
\begin{equation}\label{eq.38}
	\begin{split}
		\mathbb{E}[\bar{R}(g)]-R(g)&=\mathbb{E}[\bar{R}_n(g)-\widehat{R}_n(g)]\\
		&=\int_{S_n\in\Omega_{+}(g)}\big(\bar{R}_n(g)-R_n(g)\big)p(S_n)dS_n\\
		&+\int_{S_n\in\Omega_{-}(g)}\big(\bar{R}_n(g)-R_n(g)\big)p(S_n)dS_n\\
		&=\int_{S_n\in\Omega_{-}(g)}\big(\bar{R}_n(g)-R_n(g)\big)p(S_n)dS_n.\\
	\end{split}
\end{equation}

By the definition of $\bar{R}(g)$, it serves as an upper bound on $\widehat{R}(g)$, i.e., $\bar{R}(g) \geq \widehat{R}(g)$, which implies:
\begin{equation}\label{eq.39}
	\mathbb{E}[\bar{R}(g) - \widehat{R}(g)] \geq 0.
\end{equation}

Since the consistent correction function is Lipschitz continuous with constant $L_f = \max\{1, k\}$ and satisfies $f(0) = 0$, it follows that $\left|\widehat{R}_n(g)\right| \leq (N + 1) \tau C_w C_\ell$. Based on this, we can further bound the gap $\mathbb{E}[\bar{R}(g)] - R(g)$.
\begin{equation}\label{eq.40}
	\begin{split}
		&\mathbb{E}[\bar{R}_n(g)] - R(g) = \int_{S_n \in \Omega_-(g)} \left(\bar{R}_n(g) - \hat{R}_n(g)\right) p(S_n) dS_n \\
		&\leq \sup_{S_n \in \Omega_-(g)} \left( (\bar{R}_n(g) - \hat{R}_n(g)) \int_{S_n \in \Omega_-(g)} p(S_n) dS_n \right) \\
		&= \sup_{S_n \in \Omega_-(g)} \left( (\bar{R}_n(g) - \hat{R}_n(g)) \mathbb{P}(\Omega_-(g)) \right) \\
		&= \sup_{S_n \in \Omega_-(g)} \left( f\left(\hat{R}_n(g)\right)- \hat{R}_n(g)\right) \mathbb{P}(\Omega_-(g)) \\
		&\leq \sup_{S_n \in \Omega_-(g)} \left( L_f \left|\hat{R}_n(g)\right|  + \left|\hat{R}_n(g)\right| \right) \mathbb{P}(\Omega_-(g)) \\
		&\leq \sup_{S_n \in \Omega_-(g)} \big( (L_f + 1) (N + 1) \tau C_w C_\ell \big) \mathbb{P}(\Omega_-(g)) \\
		&=  (L_f + 1) (N + 1) \tau C_w C_\ell \exp\left(-\frac{2\alpha^2}{(Nn_b + n_c)\Delta^2}\right)
	\end{split} 
\end{equation}

We  give the following inequality:
\begin{equation}\label{eq.41}
	\begin{split}
		&\left|\bar{R}(g) - R(g)\right| 
		\leq \left|\bar{R}(g) - \mathbb{E}[\bar{R}(g)]\right| + \left|\mathbb{E}[\bar{R}(g)] - R(g)\right| \nonumber \\
		&\leq \left|\bar{R}(g) - \mathbb{E}[\bar{R}(g)]\right| \\
		&+  (L_f + 1) (N + 1) \tau C_w C_\ell \exp\left(-\frac{2\alpha^2}{(Nn_b + n_c)\Delta^2}\right)
	\end{split}
\end{equation}

Given the definition of $\bar{R}(g)$ and the Lipschitz continuity of the correction function, the change in $\bar{R}(g)$ is at most $L_\ell \Delta$. By applying McDiarmid?s inequality, we can bound the deviation $\left| \bar{R}(g) - \mathbb{E}[\bar{R}(g)] \right|$ with high probability. Specifically, with probability at least $1 - \delta$, the following inequality holds:
\begin{equation}\label{eq.42}
	\begin{split}
		\left|\bar{R}(g) - \mathbb{E}[\bar{R}(g)]\right| 
		\leq L_\ell\Delta \sqrt{\frac{\ln (2/\delta)}{2(Nn_b + n_c)}}
	\end{split}
\end{equation}

\section*{K. Proof of  Theorem 9.}\label{app9}
\textit{Proof}.
We first give the following inequalities:
\begin{equation}\label{eq.43}
	\begin{split}
		&R(\bar{g}) - R(g^*) = \left( R(\bar{g}) - \bar{R}(\bar{g}) \right) \\
		&+ \left( \bar{R}(\bar{g}) - \bar{R}(\bar{g}) \right) 
		+ \left( \bar{R}(\bar{g}) - R(\bar{g}) \right) + (R(\bar{g}) - R(g^*)) \\
		&\leq \left| R(\bar{g}) - \bar{R}(\bar{g}) \right| + \left| \bar{R}(\bar{g}) - R(\bar{g}) \right| 
		+ (R(\bar{g}) - R(g^*))
	\end{split}
\end{equation}

Then we can conclude the proof by combining the high-probability bound in Theorem 7, Theorem 4 and union bound. With probability at least $1 - \delta$, the following inequality holds:
\begin{equation}\label{eq.44}
	\begin{split}
		&R(\bar{g}) - R(g^*) \leq \left| R(\bar{g}) - \bar{R}(\bar{g}) \right| \\
		&+ \bar{R}(\bar{g}) - \bar{R}(\hat{g})
		+ \left| \bar{R}(\hat{g}) - R(\hat{g}) \right| + (R(\bar{g}) - R(g^*)) \\
		&\leq 2 L_\ell\Delta \sqrt{\frac{\ln (2/\delta)}{2n}} \\
		&+ 2 \mathcal{O}(1) \cdot \exp\left(-\frac{2\alpha^2}{n\Delta^2}\right) 
		+ K_{n}\frac{1}{\sqrt{n_b}}+K_{u}\frac{1}{\sqrt{n_{u}}}
	\end{split}
\end{equation}

\bibliographystyle{ieeetr}
\bibliography{ref}

\begin{IEEEbiography}[{\includegraphics[width=1in,height=1.25in,clip,keepaspectratio]{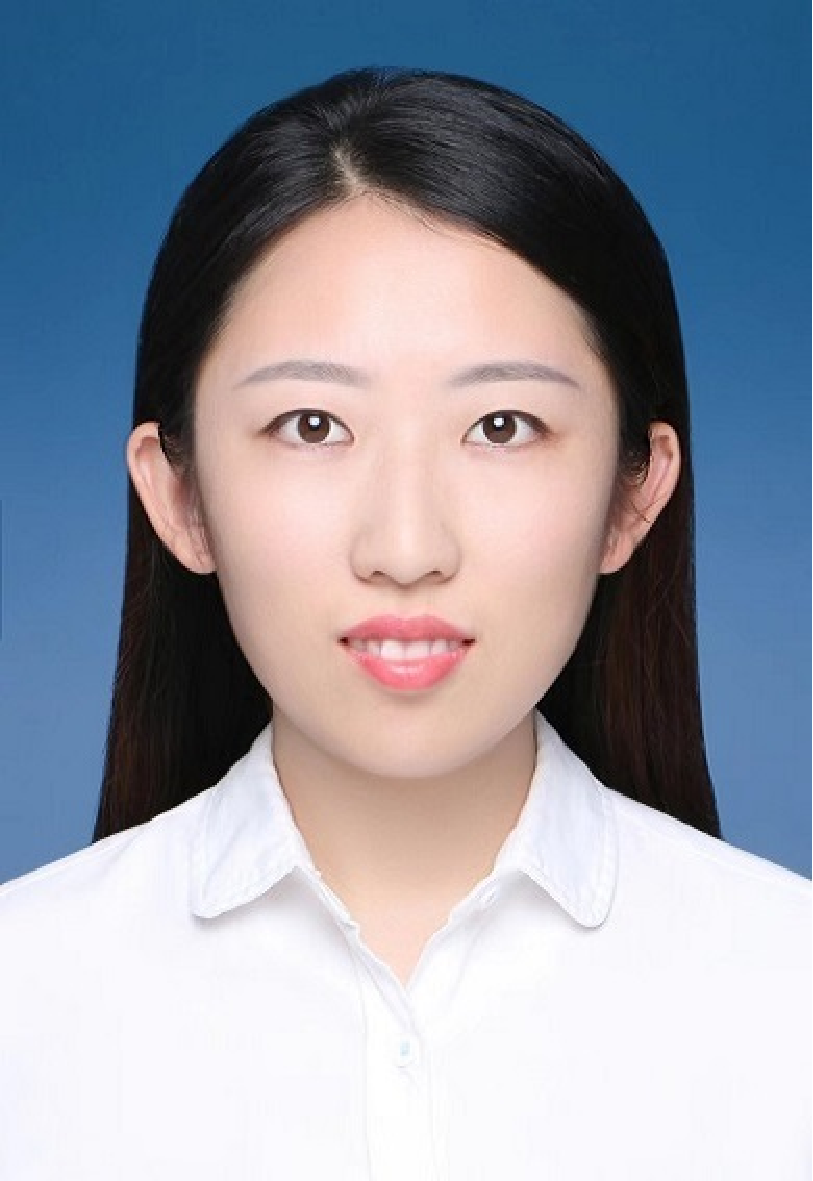}}]{Shuying Huang}
	obtained her bachelor's degree in Automation from Yantai University, Yantai, China in 2019 and is currently pursuing  Ph.D. degree in Control Engineering from Yanshan University, Qinhuangdao, China. Her research interests are in weakly supervised machine learning.
\end{IEEEbiography}

\begin{IEEEbiography}[{\includegraphics[width=1in,height=1.25in,clip,keepaspectratio]{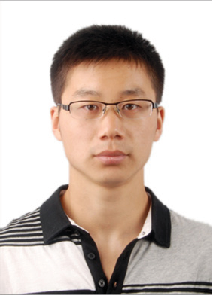}}]{Junpeng Li}
	received the B.Sc. degree in Biomedical Engineering and the Ph.D. degree in Control science and Engineering, both from Yanshan University, China, in 2010 and 2016, respectively. Currently, he is Full Professor in the Department of Automation at Yanshan University, China. His current research interests include system modeling, machine learning and intelligent optimization.
\end{IEEEbiography}

\begin{IEEEbiography}[{\includegraphics[width=1in,height=1.25in,clip,keepaspectratio]{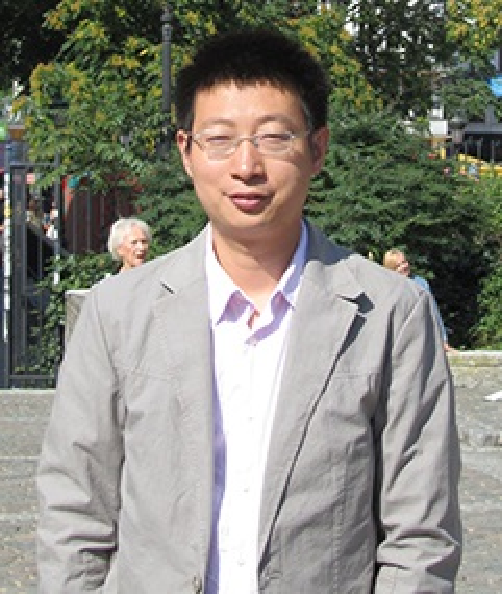}}]{Changchun Hua} received the Ph.D degree in electrical engineering from
	Yanshan University, Qinhuangdao, China, in 2005. He was a research Fellow in
	National University of Singapore from 2006 to 2007. From 2007 to 2009, he
	worked in Carleton University, Canada, funded by Province of Ontario Ministry
	of Research and Innovation Program. From 2009 to 2010, he worked in University
	of Duisburg-Essen, Germany, funded by Alexander von Humboldt Foundation. Now
	he is a full Professor in Yanshan University, China. He is the author or
	coauthor of more than 80 papers in mathematical, technical journals, and
	conferences. He has been involved in more than 10 projects supported by the
	National Natural Science Foundation of China, the National Education Committee
	Foundation of China, and other important foundations. His research interests
	are in nonlinear control systems, control systems design over network,
	teleoperation systems and intelligent control.
\end{IEEEbiography}

\begin{IEEEbiography}[{\includegraphics[width=1in,height=1.25in,clip,keepaspectratio]{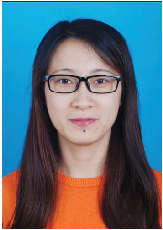}}]{Yana Yang}
	received her Ph.D. degree in Electrical Engineering from Yanshan University, Qinhuangdao, China, in 2017. Now she is currently a  Associate Professor of Department of Automation, Yanshan University, Qinhuangdao 066004, China. She is the author or coauthor of more than 20 papers in mathematical, technical journals, and conferences. Her research interests are in nonlinear teleoperation system control, nonlinear control systems, robot system control and sliding mode control.
\end{IEEEbiography}

\end{document}